\documentclass[sigconf,natbib,screen=true,review=false]{acmart}
\PassOptionsToPackage{dvipsnames}{xcolor}

\usepackage{dsfont}
\usepackage{acronym}
\usepackage[noend]{algorithmic}
\usepackage[ruled,vlined,linesnumbered]{algorithm2e}
\usepackage{bbm}
\usepackage{beramono}
\usepackage{bm}
\usepackage{booktabs}
\usepackage[skip=0pt]{caption}
\usepackage{cleveref}
\usepackage[T1]{fontenc}

\usepackage{graphicx}
\usepackage{hyperref}
\usepackage{listings}
\usepackage{multirow}
\usepackage{subcaption}
\usepackage{tabularx}
\usepackage[dvipsnames]{xcolor}
\usepackage[inline]{enumitem}
\usepackage{numprint}
\usepackage[normalem]{ulem}

\usepackage{balance}
\usepackage{mathrsfs}
\usepackage{mathtools}
\usepackage{xspace}

\usepackage{tikz}
\usetikzlibrary{positioning, fadings,patterns,fit}
\usetikzlibrary{arrows}

\makeatletter
\newif\ifAC@uppercase@first%
\def\Aclp#1{\AC@uppercase@firsttrue\aclp{#1}\AC@uppercase@firstfalse}%
\def\AC@aclp#1{%
  \ifcsname fn@#1@PL\endcsname%
    \ifAC@uppercase@first%
      \expandafter\expandafter\expandafter\MakeUppercase\csname fn@#1@PL\endcsname%
    \else%
      \csname fn@#1@PL\endcsname%
    \fi%
  \else%
    \AC@acl{#1}s%
  \fi%
}%
\def\Acp#1{\AC@uppercase@firsttrue\acp{#1}\AC@uppercase@firstfalse}%
\def\AC@acp#1{%
  \ifcsname fn@#1@PL\endcsname%
    \ifAC@uppercase@first%
      \expandafter\expandafter\expandafter\MakeUppercase\csname fn@#1@PL\endcsname%
    \else%
      \csname fn@#1@PL\endcsname%
    \fi%
  \else%
    \AC@ac{#1}s%
  \fi%
}%
\def\Acfp#1{\AC@uppercase@firsttrue\acfp{#1}\AC@uppercase@firstfalse}%
\def\AC@acfp#1{%
  \ifcsname fn@#1@PL\endcsname%
    \ifAC@uppercase@first%
      \expandafter\expandafter\expandafter\MakeUppercase\csname fn@#1@PL\endcsname%
    \else%
      \csname fn@#1@PL\endcsname%
    \fi%
  \else%
    \AC@acf{#1}s%
  \fi%
}%
\def\Acsp#1{\AC@uppercase@firsttrue\acsp{#1}\AC@uppercase@firstfalse}%
\def\AC@acsp#1{%
  \ifcsname fn@#1@PL\endcsname%
    \ifAC@uppercase@first%
      \expandafter\expandafter\expandafter\MakeUppercase\csname fn@#1@PL\endcsname%
    \else%
      \csname fn@#1@PL\endcsname%
    \fi%
  \else%
    \AC@acs{#1}s%
  \fi%
}%
\edef\AC@uppercase@write{\string\ifAC@uppercase@first\string\expandafter\string\MakeUppercase\string\fi\space}%
\def\AC@acrodef#1[#2]#3{%
  \@bsphack%
  \protected@write\@auxout{}{%
    \string\newacro{#1}[#2]{\AC@uppercase@write #3}%
  }\@esphack%
}%
\def\Acl#1{\AC@uppercase@firsttrue\acl{#1}\AC@uppercase@firstfalse}
\def\Acf#1{\AC@uppercase@firsttrue\acf{#1}\AC@uppercase@firstfalse}
\def\Acfi#1{\AC@uppercase@firsttrue\acfi{#1}\AC@uppercase@firstfalse}
\def\Ac#1{\AC@uppercase@firsttrue\ac{#1}\AC@uppercase@firstfalse}
\def\Acs#1{\AC@uppercase@firsttrue\acs{#1}\AC@uppercase@firstfalse}
\robustify\Aclp
\robustify\Acfp
\robustify\Acp
\robustify\Acsp
\robustify\Acl
\robustify\Acf
\robustify\Acfi
\robustify\Ac
\robustify\Acs
\makeatother

\def\AC@@acro#1[#2]#3{%
  \ifAC@nolist%
  \else%
  \ifAC@printonlyused%
    \expandafter\ifx\csname acused@#1\endcsname\AC@used%
       \item[\protect\AC@hypertarget{#1}{\acsfont{#2}}] #3%
          \ifAC@withpage%
            \expandafter\ifx\csname r@acro:#1\endcsname\relax%
               \PackageInfo{acronym}{%
                 Acronym #1 used in text but not spelled out in
                 full in text}%
            \else%
               \dotfill\pageref{acro:#1}%
            \fi\\%
          \fi%
    \fi%
 \else%
    \item[\protect\AC@hypertarget{#1}{\acsfont{#2}}] #3%
 \fi%
 \fi%
 \begingroup
    \def\acroextra##1{}%
    \@bsphack
    \protected@write\@auxout{}%
       {\string\newacro{#1}[\string\AC@hyperlink{#1}{#2}]{\AC@uppercase@write #3}}%
    \@esphack
  \endgroup}

\definecolor{ali}{RGB}{0, 150, 0}
\definecolor{mdr}{RGB}{255, 127, 0}

\acrodef{CDC}{Cosine-Distance Criterion}
\newcommand{\ourmethod}{\ac{CDC}\xspace}
\newcommand{\ourmethodtitle}{\acf{CDC}\xspace}

\newcommand{\myparagraph}[1]{\smallskip\emph{\bfseries #1.}}
\newenvironment{blockquote}{%
  \par%
  \medskip
  \leftskip=2em\rightskip=2em%
  \noindent\ignorespaces}{%
  \par\medskip}

\newcommand{\mExpect}[2]{\ensuremath{\mathbb{E}_{#1}\left[#2\right]}}

\newcommand{\moped}{overparameterized\xspace}
\newcommand{\mop}{overparameterization\xspace}
\newcommand{\mOped}{Overparameterized\xspace}
\newcommand{\mOp}{Overparameterization\xspace}

\acrodef{IR}{information retrieval}
\acrodef{EM}{Expectation-Maximization}
\acrodef{rbEM}{regression-based EM}
\acrodef{LTR}{learning to rank}
\acrodef{SERP}{Search Engine Results Page}
\acrodef{CLTR}{counterfactual learning to rank}
\acrodef{OLTR}{Online Learning to Rank}
\acrodef{IPS}{inverse propensity scoring}
\acrodef{PBM}{Position-Based Model}
\acrodef{CM}{Cascade Model}
\acrodef{ARP}{Average Relevance Position}
\acrodef{DCG}{Discounted Cumulative Gain}
\acrodef{DLA}{Dual Learning Algorithm}
\acrodef{MNAR}{Missing-Not-At-Random}
\acrodef{mPMM}{modified Poisson Mixture Model}
\acrodef{DCM}{Dependent Click Model}
\acrodef{UBM}{User Browsing Model}
\acrodef{CCM}{Click Chain Model}
\acrodef{PMM}{Poisson Mixture Model}
\acrodef{GMM}{Gaussian Mixture Model}
\acrodef{BMM}{Binomial Mixture Model}
\acrodef{CTR}{click-through rate}
\acrodef{AC}{affine correction}
\acrodef{i.i.d.}{independent and identically distributed}
\acrodef{CLT}{Central Limit Theorem}
\acrodef{NDCG}{Normalized Discounted Cumulative Gain}
\acrodef{MAP}{Mean Average Precision}
\acrodef{LOO}{leave one out}
\acrodef{SGD}{stochastic gradient descent}
\acrodef{GD}{gradient descent}
\acrodef{CV}{cross validation}
\acrodef{NN}{neural network}

\theoremstyle{definition}

\theoremstyle{definition}

\allowdisplaybreaks

\makeatletter
\renewcommand\paragraph{\@startsection{paragraph}{4}{\parindent}%
  {-.1\baselineskip \@plus -1\p@ \@minus -.1\p@}%
  {-2\p@}%
  {\ACM@NRadjust{\@parfont\@adddotafter}}}
\makeatother  

\newcommand{\ROne}[1]{\textcolor{teal}{\textbf{[R1]}}}
\newcommand{\RTwo}[1]{\textcolor{blue}{\textbf{[R2]}}}
\newcommand{\RThree}[1]{\textcolor{cyan}{\textbf{[R3]}}}

\copyrightyear{2022} 
\acmYear{2022} 
\setcopyright{acmlicensed}\acmConference[CIKM '22]{Proceedings of the 31st ACM International Conference on Information and Knowledge Management}{October 17--21, 2022}{Atlanta, GA, USA}
\acmBooktitle{Proceedings of the 31st ACM International Conference on Information and Knowledge Management (CIKM '22), October 17--21, 2022, Atlanta, GA, USA}
\acmPrice{15.00}
\acmDOI{10.1145/3511808.3557366}
\acmISBN{978-1-4503-9236-5/22/10}

\author{Ali Vardasbi}
\orcid{0000-0002-9342-5272}
\affiliation{%
  \institution{University of Amsterdam}
  \city{Amsterdam}
  \country{The Netherlands}
}
\email{a.vardasbi@uva.nl}

\author{Maarten de Rijke}
\orcid{0000-0002-1086-0202}
\affiliation{%
  \institution{University of Amsterdam}
  \city{Amsterdam}
  \country{The Netherlands}
}
\email{m.derijke@uva.nl}

\author{Mostafa Dehghani}
\orcid{0000-0002-9772-1095}
\affiliation{%
  \institution{Google Brain}
  \city{Amsterdam}
  \country{The Netherlands}
}
\email{dehghani@google.com}

\title[Intersection of Parallels as an Early Stopping Criterion]{Intersection of Parallels as an Early Stopping Criterion}

\allowdisplaybreaks

\begin{document}

\begin{abstract}
	
    A common way to avoid overfitting in supervised learning is early stopping, where a held-out set is used for iterative evaluation during training to find a sweet spot in the number of training steps that gives maximum generalization. 
	However, such a method requires a disjoint validation set, thus part of the labeled data from the training set is usually left out for this purpose, which is not ideal when training data is scarce.
	Furthermore, when the training labels are noisy, the performance of the model over a validation set may not be an accurate proxy for generalization.
	In this paper, we propose a method to spot an early stopping point in the training iterations of an \moped \ac{NN} without the need for a validation set.
	We first show that in the \moped regime the randomly initialized weights of a linear model converge to the same direction during training.
	Using this result, we propose to train \emph{two parallel instances} of a linear model, initialized with different random seeds, and use their \emph{intersection} as a signal to detect overfitting.
	In order to detect intersection, we use the cosine distance between the weights of the parallel models during training iterations.
	Noticing that the final layer of a \ac{NN} is a linear map of pre-last layer activations to output logits, we build on our criterion for linear models and propose an extension to multi-layer networks, using the new notion of \emph{counterfactual weights}.
	We conduct experiments on two areas that early stopping has noticeable impact on preventing overfitting of a \ac{NN}:
	\begin{enumerate*}[label=(\roman*)]
		\item learning from noisy labels; and
		\item \acl{LTR} in \acl{IR}.
	\end{enumerate*}
	Our experiments on four widely used datasets confirm the effectiveness of our method for generalization.
	For a wide range of learning rates, our method, called \ourmethodtitle, leads to better generalization on average than all the methods that we compare against in almost all of the tested cases.
\end{abstract}



\maketitle

\acresetall



\section{Introduction}
\label{sec:intro} 

Modern \moped neural networks can easily overfit to the training data, even when the data is noisy~\citep{li2020gradient}. 
Collecting up-to-date and large-scale high accuracy labeled data is however expensive, time-consuming, and in many cases not feasible.
The main question we address in this work is ``how can we improve generalizability of \moped neural networks in low data regimes and noisy labeled data?''

\myparagraph{Early stopping}
Early stopping is one of the fundamental techniques for generalization in \acp{NN} with iterative training~\citep{montavon2012neural, goodfellow2016deep}.
The common practice for early stopping is to monitor model performance on a held-out set, called validation set, and quit training if the validation performance degrades or has not improved for several iterations.
In such settings, the assumption is that model performance over the held-out data, which has not been seen by the model during training, is an unbiased proxy for the test performance.
This poses a trade-off for choosing the size of the validation set: A small validation set would suffer from high variance in estimating the test performance, whereas a large validation set would reduce the size of the training set and possibly hurt the model generalizability.
Additionally, it is known that \moped \acp{NN} can easily overfit noisy labeled datasets without early stopping~\citep{li2020gradient}.
However, as noted in~\citep{forouzesh2021disparity}, when training on a noisy labeled dataset, relying on an accurately labeled validation set for early stopping is not realistic. 

Ideally, then, we would like to have a criterion for early stopping that does not rely on the existence of a validation set.
Some validation set independent criteria have been proposed in the literature based on gradients~\citep{mahsereci2017early, forouzesh2021disparity} or \ac{LOO} interpolation~\citep{bonet2021channel}.

\myparagraph{\mOp}
Despite the positive generalizability results of \mop that state that \moped models are less dependent on regularization~\citep{zhang2016understanding, belkin2019reconciling}, early stopping has been shown to still be helpful in \moped models~\citep{kuzborskij2021nonparametric}, especially in the presence of label noise~\citep{li2020gradient}.

\myparagraph{A proposal for early stopping}
In this work, we propose a new criterion for early stopping \acp{NN} with iterative optimization methods such as (stochastic) gradient descent, that is particularly helpful in low data regimes and noisy labeled datasets.
We start with linear models and then describe how to extend our method to be effective for non-linear multi-layer networks as well.
Our early stopping criterion is based on our experimental observation that  different instances of an \moped{} linear model, initialized with different random seeds, tend to converge to a similar solution on the fitness surface during training.
\begin{figure}[t]
    \centering
    \begin{tabular}[]{@{}l@{~}c@{~}r@{~}c}

        \rotatebox[origin=lt]{90}{\hspace{2.2em} \small Accuracy}
        &
        \includegraphics[scale=0.32]{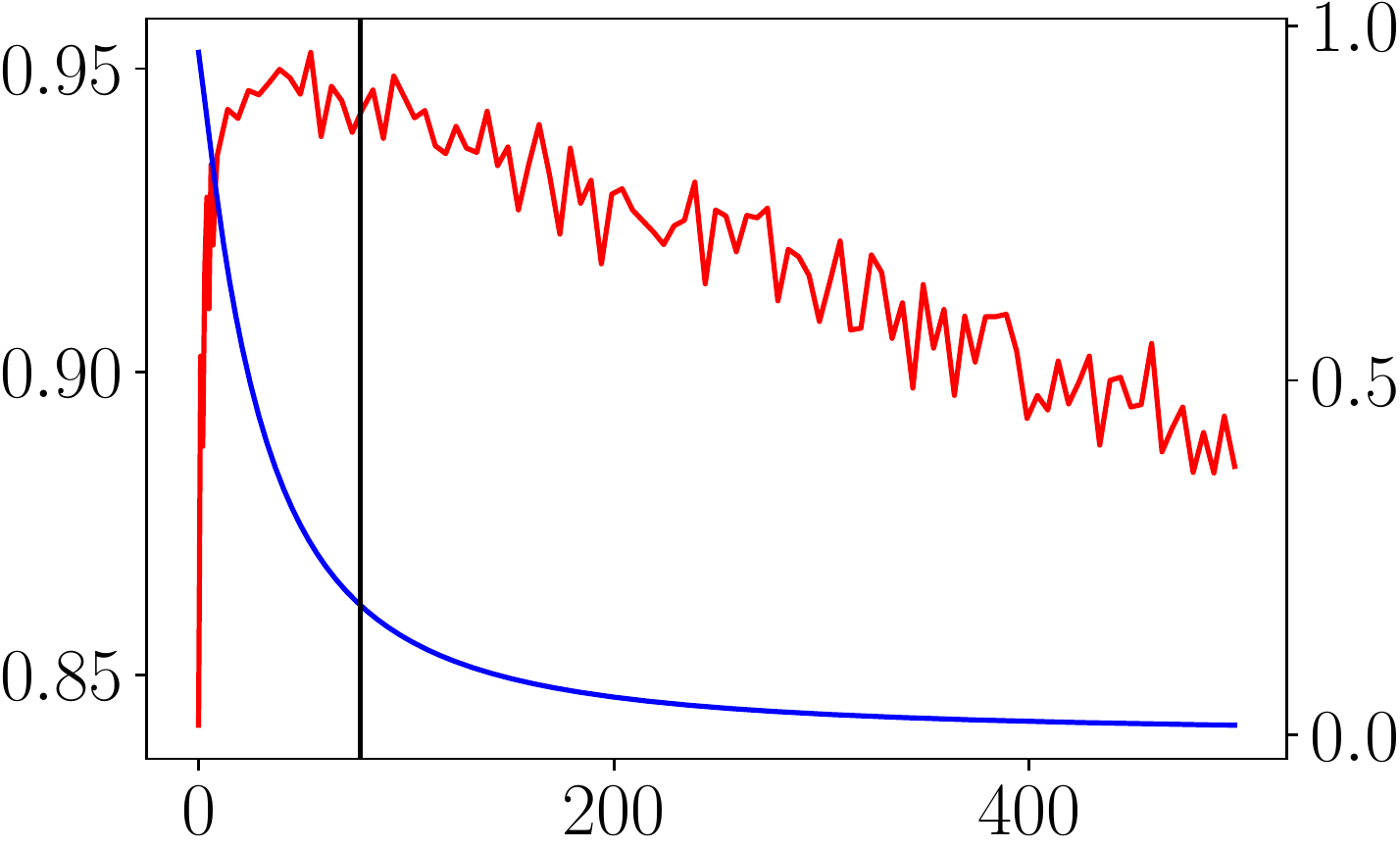}
        &
        \rotatebox[origin=lt]{90}{\hspace{1em} \small Cosine distance}
        &
        \includegraphics[scale=0.40]{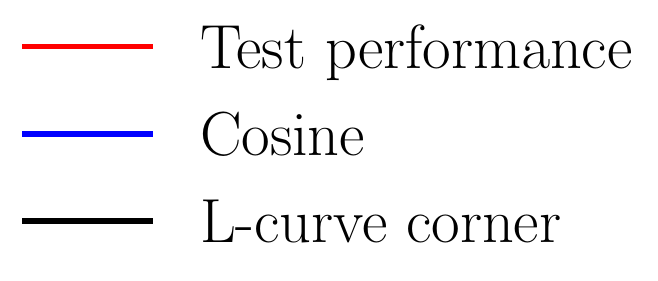}  \\
        & \small Epochs
        
    \end{tabular}
    \caption{Convergence of cosine distance of parallel instances coincides with overfitting.}
    \label{fig:lcurve}
\end{figure}
Fig.~\ref{fig:lcurve} plots the cosine distance between the weight vectors of two linear networks as well as test performance during the training. 
First, we observe that the cosine distance forms an L-curve.
We indicate the corner of this L-curve by a vertical black line.
We hypothesize that when the cosine distance starts to converge to zero, i.e., the two parallel instances intersect, the models have already learned the ``generalizable'' patterns for solving the task and after that point they start picking up spurious features that only represent the training set.
This is observable in Fig.~\ref{fig:lcurve}, where the test performance continues to decrease after the L-curve corner.
Our goal is to find the transition point between the two extreme cases: 
\begin{enumerate*}[label=(\roman*)]
\item The first training iterations where the parallel instances are far from each other. This case is captured by weight vectors that are randomly directed and are almost surely orthogonal.
\item The converged iterations where the parallel instances become very close to each other, i.e., intersect. This case is captured by the weight vectors that are aligned with the direction of zero training loss.
\end{enumerate*}
To this end, we measure the cosine distance between the weight vectors of parallel models at each iteration.
Due to the two extreme cases mentioned above, the cosine distances with respect to training iterations form an L-curve (Fig.~\ref{fig:lcurve}).
We use two simple methods, namely the maximum curvature and a threshold-based criterion, to detect the transition point, and output the corresponding iteration as the early stopping point.
Our experiments on a diverse set of datasets on different data modalities show that the transition point obtained as above for the early stopping iteration leads to better generalization results than existing methods.

\myparagraph{Going beyond linear models}
Our observation of converging parallel instances only holds for linear models where the weight vector is used to map a fixed input matrix to a fixed target vector.
For networks with more than one layer and non-linear activation functions, however, the weight vector of the last layer maps the activation matrix of the previous layer to the target vector.
So for two different randomly initialized networks $\mathscr{N}_1$ and $\mathscr{N}_2$, last layer activations, $A_1$ and $A_2$, may be in completely different spaces and not directly comparable.
Consequently, the described L-curve criterion cannot be used as-is for the linear mapping in the last layer of networks with multiple layers.
To remedy this, we first answer the counterfactual question of: \emph{what would be the weight vector of $\mathscr{N}_2$ if its last layer activation $A_2$ was to be replaced with $A_1$ from $\mathscr{N}_1$?}
Using a linear algebra trick, we find an answer for this counterfactual question, and as the settings have been made similar to the linear case, we observe a similar L-curve trend in the cosine distance between the weight vector of $\mathscr{N}_1$ and the counterfactual weight vector of $\mathscr{N}_2$.
Our experiments on different datasets and with a wide range of learning rates confirm the effectiveness of this method on multi-layer networks.


Our contributions in this paper are:
\begin{enumerate}[label=(\arabic*),leftmargin=*,nosep]
	\item We empirically show that for \moped linear models, the randomly initialized weight vectors, when trained by an iterative method such as \ac{SGD}, converge to the same solution. 
	\item We use the above finding on linear models to detect an early stopping point without a validation set by measuring the cosine distance between the weight vectors of two parallel instances of a linear model, initialized with different random seeds.
	We call our method \acfi{CDC}. 
	\item Using the counterfactual weight vector of the last layer of one of the instances, we extend our proposed cosine distance criterion to multi-layer networks.
	\item We experimentally verify the generalization effectiveness of \ourmethod{} on two widely used computer vision datasets with noisy labels as well as two widely used \ac{LTR} datasets.
	Particularly, we show that \ourmethod leads to better average test performance, i.e., better generalization, than the methods against which we compare across a wide range of learning rates.
\end{enumerate}

\subsection{Notation}
\label{sec:multilayer:notations}
Let $X_{N\times d}=[x_1~x_2~\cdots~x_N]^T$ denote the training data consisting of $x_i\in \mathbb{R}^d$ as feature vectors and let $Y_{N \times 1} =[y_1~y_2~\cdots~y_N]^T$ denote the corresponding labels.
For a one-layer network, i.e., a linear model, we use a (non-trainable) featurization function $\Phi:\mathbb{R}^d \rightarrow \mathbb{R}^D$ with $D>N$ to have an \moped model.
We use $A_{N\times D}=[\Phi(x_1)~\cdots~\Phi(x_N)]^T$ for the featurized inputs.
Training a linear model means finding a weight vector $W=[w_1, w_2, \ldots, w_D]^T$ satisfying:
\begin{equation}
    \label{eq:linearmodel}
    A\cdot W = Y.
\end{equation}
\noindent
For a $m$-layer network, layer $i$ can be written as:
\begin{equation}
    \label{eq:multilayernetwork}
    A_{i+1} = \sigma \left(A_i \cdot W_i\right)
\end{equation}
\noindent
where $\sigma$ is the activation function, and $A_0=X$.\footnote{Without loss of generality we assume the bias is zero. We can always concatenate a single $1$ to the end of feature vector $x$ to model the bias.}
Focusing on the last layer, training such a network means finding parameters leading to activation matrix $A_m$ and a weight vector $W_m$ satisfying:
\begin{equation}
    \label{eq:multilayernetwork}
    A_m \cdot W_m = Y.
\end{equation}
\noindent
Here, as opposed to Eq.~\eqref{eq:linearmodel} where $A$ only depends on the input data, $A_m$ depends on the trainable parameters.
This means that, for different initializations, $A_m$ in Eq.~\eqref{eq:multilayernetwork} will be different, while $A$ in Eq.~\eqref{eq:linearmodel} is fixed.
In what follows, when no confusion is possible, we drop the subscripts and write $A$ and $W$ for the pre-last layer activation matrix and last layer weight vector of a multi-layer network, respectively.
For both linear and multi-layer networks, we use subscripts to differentiate between instances that are initialized with different random seeds.
We use $A^+$ for the Moore-Penrose inverse of matrix $A$.

Finally, we show \ac{GD} iterations by parenthesized superscripts.
For example, iteration $t$ of a multi-layer network satisfies:
\begin{equation}
    A^{(t)}\cdot W^{(t)} = \Tilde{Y}^{(t)},
\end{equation}
\noindent
where $\Tilde{Y}$ is the model prediction.
Iteration $0$ stands for the initialization values.

\section{Related Work}
\label{sec:related}
{\emph{\bfseries Early stopping.}}
The idea of early stopping in machine learning is based on the assumption that in the process of training iterations of a model, the model first learns the general relation between the inputs and labels, and then, gradually overfits to the training samples~\citep{montavon2012neural}.
Various studies investigate the effect of early stopping or propose rules to identify early stopping points, especially with the gradient descent algorithm~\citep{yao2007early, kuzborskij2021nonparametric}.
A common practice for early stopping is to hold out part of the training data, usually referred to as the validation set, and train the model on the remaining data.
The validation set is used to periodically evaluate the model during training, and the validation performance is considered as a generalization proxy.

As noted in previous work on early stopping~\citep{mahsereci2017early, bonet2021channel}, using a validation set for early stopping has some drawbacks, especially in the low data regime.
As the validation performance is going to be used as a proxy for generalization, its size should not be small.
On the other hand, when the annotated data is already small, consuming a considerable part of it for the validation set, reduces the size of the training data and can lead to degraded performance.
Relying on the validation set is also not desired for noisy labeled data where the performance on a noisy labeled validation set may not be a good proxy for the actual performance on unseen data~\citep{forouzesh2021disparity}.

These limitations have led some researchers to propose early stopping rules without a validation set.
\citet{mahsereci2017early}, building on~\citep{duvenaud2016early}, use gradients to identify the early stopping point.
The idea is when the gradients become small, it is a signal indicating the model has learned the general structure, and it is starting to overfit to the train data.
Similarly,~\citet{forouzesh2021disparity} use the gradients for an early stopping criterion.
Instead of the size of the gradients, they measure the disparity of gradients, i.e., the effect of a gradient step on one mini-batch on another one, and decide to stop early when a number of training iterations with increased disparities are observed.
Finally,~\citet{bonet2021channel} use \ac{LOO} interpolation to characterize generalization and propose to stop the training if for a pre-defined number of iterations the risk estimated from \ac{LOO} interpolations is increased.
As their method uses \ac{LOO} samples, it can be thought of as an efficient modified \ac{CV} with as many folds as the size of training data.

Our \ourmethod method differs from previous work in that it uses two parallel instances of a model to detect early stopping.
We show experimentally that the change of parameter vector direction during training of one network does not provide a helpful signal for overfitting.
Furthermore, instead of monitoring the gradients of the parameters, we track the parameters themselves.
Parameter values at each iteration depend on their initial values, their gradients, and the learning rate.
As such, our method can better adapt to different learning rates compared to previous work that uses gradients for their early stopping rule.
We will verify this intuition in our experiments.
Lastly, \ourmethod is agnostic to the optimization method used, so it is the same for gradient descent and stochastic gradient descent methods (unlike~\citep{mahsereci2017early}) and does not depend on the mini-batch size or the subset size of mini-batches to approximate gradient discrepancy (unlike~\citep{forouzesh2021disparity}). 

\citet{li2020gradient} analyze the robustness of neural networks to noisy labels, when trained with gradient descent.
They show that \moped networks with early stopping, where the parameters are still close to their initial values, can robustly learn the correct labels and ignore the noisy ones.
But after many more iterations, where the model goes far from its initialization, overfitting to the noisy labels occurs.
The intuition of our \ourmethod method is very close to that work in the sense that we stop training as soon as the parameters become far from their initial value.
The difference is that we use two parallel instances of a model to estimate the point with lowest generalization risk.
Using two instances allows us to spot the desired point even with large learning rates, where the model overfits soon and looking at the distance of one model to its initialization point is not enough (see Sec.~\ref{subsec:ourmethod}).

\myparagraph{\mOped models}
A model is said to be \emph{\moped} when its number of trainable parameters is bigger than the number of training samples.
Modern machine learning models are usually highly \moped with good generalization properties.
In order to better understand \moped neural networks, various studies analyze two-layer networks and examine their generalization characteristics~\citep{soltanolkotabi2018theoretical, allen2019learning, li2018learning, zhou2021local}.
\mOp is essential for this work, because only in this case the gradient descent converges to the solution with minimum $\ell_2$ distance from the initialization point~\citep{engl1996regularization}.
Since we base our early stopping rule on the convergence of weights of linear models, our rule only works with \moped models.
In modern machine learning, this is not a concern, as almost all the leading models are \moped.


A number of studies analyze \moped linear models, or two-layer wide networks with only the last layer trainable~\citep{belkin2019reconciling, nagarajan2019generalization}.
\citet{muthukumar2021classification} and~\citet{kini2020analytic} analyze the impact of loss on generalization of linear \moped models.
\citet{montanari2020interpolation} characterize the generalization error of minimum-$\ell_2$ norm solution of linear models.
There are also studies such as~\citep{arora2018convergence, soudry2018implicit, deng2019model} that analyze gradient descent on linearly separable data with \moped linear models.


\myparagraph{L-curve corner detection}
Detecting the corner of an L-curve is a popular regularization method for solving systems of linear
equations with ill-conditioned matrices~\citep{hansen1998rank}.
There are several advanced methods to solve this problem, \citep[e.g.,][]{castellanos2002triangle, kim2007regularization}, but for the purpose of our work, we adopt two basic methods:
\begin{enumerate*}[label=(\roman*)]
    \item The maximum curvature point~\citep{hansen1998rank}; and
    \item A fixed threshold on the cosine distance as an indication of the transition from generalization to overfitting.
\end{enumerate*}
We show with extensive experiments that these two basic methods work fine in detecting the transition point for early stoping.


\section{\acl{CDC}}
\label{sec:method} 
The \acf{CDC} for early stopping is based on the convergence of weight vectors of two instances of a linear model that are initialized with different random seeds.
In this section, we first discuss this convergence behavior and then show how it can be used for early stopping in a linear model.
Finally, we present an extension of~\ourmethod to multi-layer networks.

\subsection{Convergence of Weights in Linear Model}
\label{subsec:convergence}
\begin{figure}[t]
    \centering
    \begin{tabular}{@{}l@{~}c}

        \rotatebox[origin=lt]{90}{\hspace{-0.1em} \small Squared $\ell_2$ norm}
        &
        \includegraphics[scale=0.27]{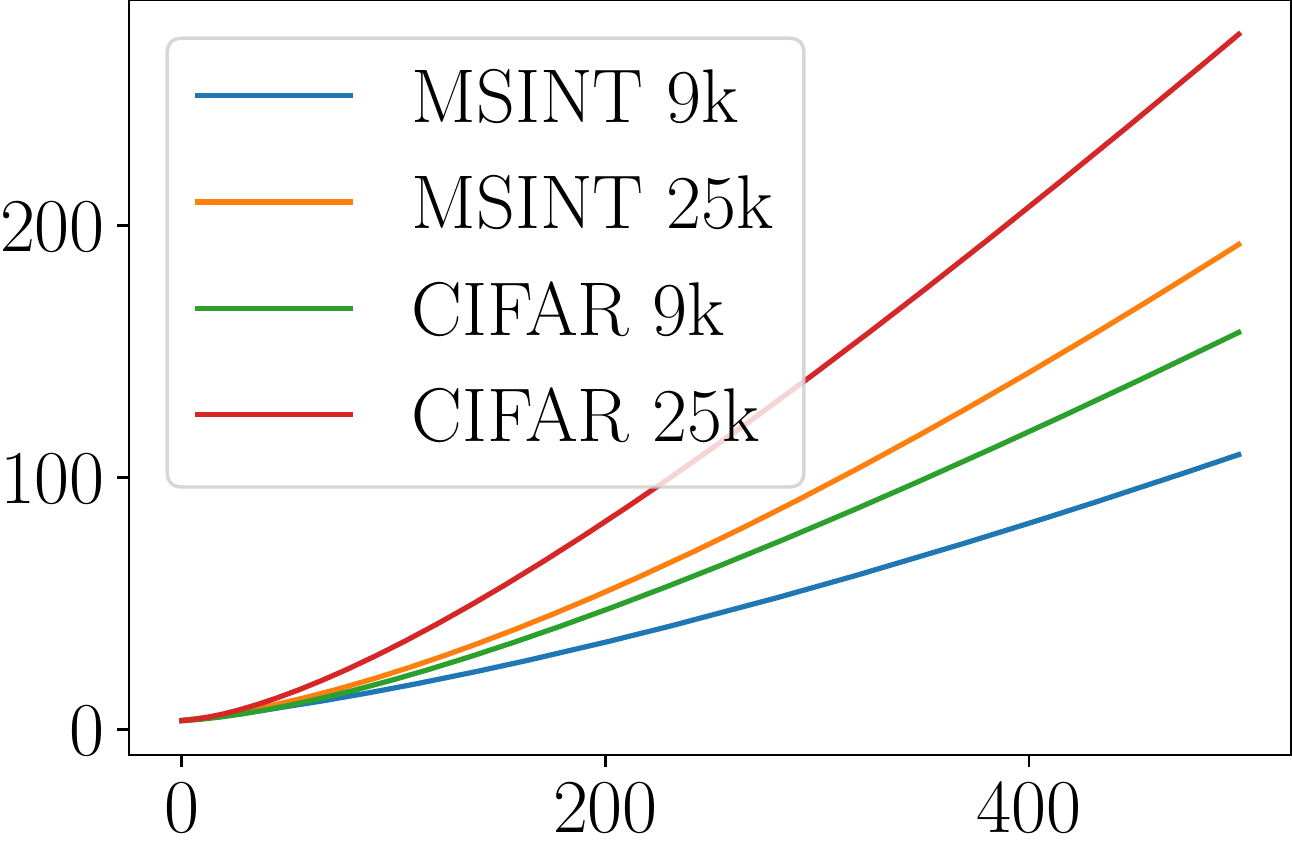} \\
        & \hspace{0.1em} \small Epochs
        
    \end{tabular}
    \caption{The growth of the weight vectors $\ell_2$ norms for MNIST and CIFAR datasets and hidden layer widths of $9$k and $25$k.}
    \label{fig:norms}
\end{figure}

In an \moped system, there are infinite directions of weight vectors that minimize the training loss.
In the case of the RMSE loss, it can be shown that gradient descent converges to the solution with minimum $\ell_2$ distance from the initialization point~\citep{engl1996regularization}.
The common practice in current neural models is to use the Xavier initialization~\citep{glorot2010understanding} 
that gives rise to
    $\mExpect{}{|W^{(0)}|^2} = \frac{1}{3}$.
\noindent
As proved in~\citep{li2020gradient} for a wide class of datasets, including noisy labeled datasets, network weights should converge to large values so that the network is able to distinguish inputs with small difference, but different labels.
Our empirical results confirm this for all of our tested datasets.
For example, Fig.~\ref{fig:norms} illustrates the growth of $\ell_2$-norms of linear models on MNIST and CIFAR datasets with different widths for the hidden layer ($9$k and $25$k).
Consequently, the $\ell_2$-norm of the initial weights becomes negligible compared to the $\ell_2$-norm in the higher iterations.
This means that the $\ell_2$-norm of the converged weights becomes dominant and the $\ell_2$ distance from the initialization point can be approximated by the $\ell_2$-norm itself.
In other words,\looseness=-1 
\begin{blockquote}
    Given two instances of an \moped linear model that are initialized using Xavier initialization, but with different random seeds, training them using  \emph{\ac{SGD} leads to similar solutions on the fitness surface.}
\end{blockquote}
%

\begin{figure}[t!]
    \centering
    \begin{tabular}[]{l@{~}ccl@{~}c}

        \rotatebox[origin=lt]{90}{\hspace{0.1em} \small Cosine distance}
        &
        \includegraphics[scale=0.27]{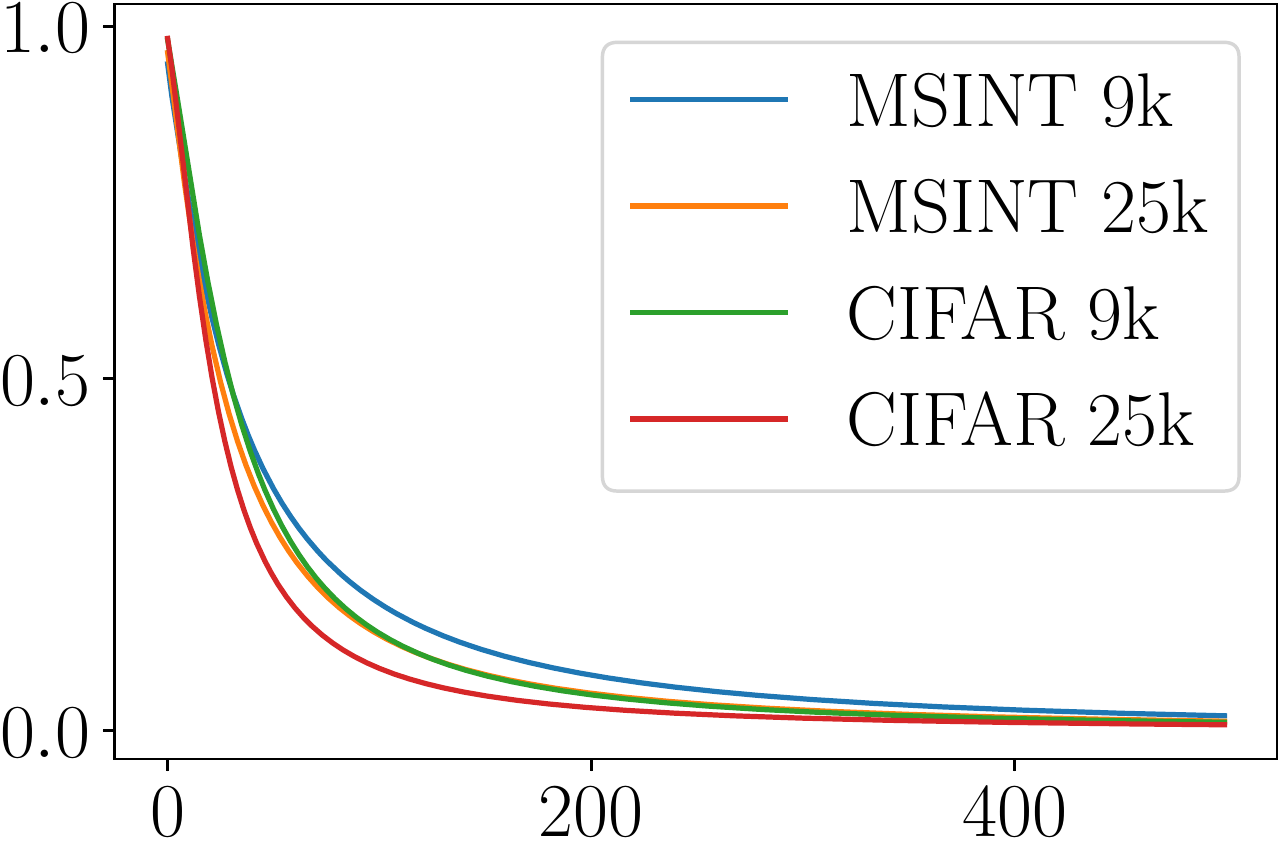} 
        & & 
        \rotatebox[origin=lt]{90}{\hspace{-0.3em} \small Euclidean distance}
        &
        \includegraphics[scale=0.27]{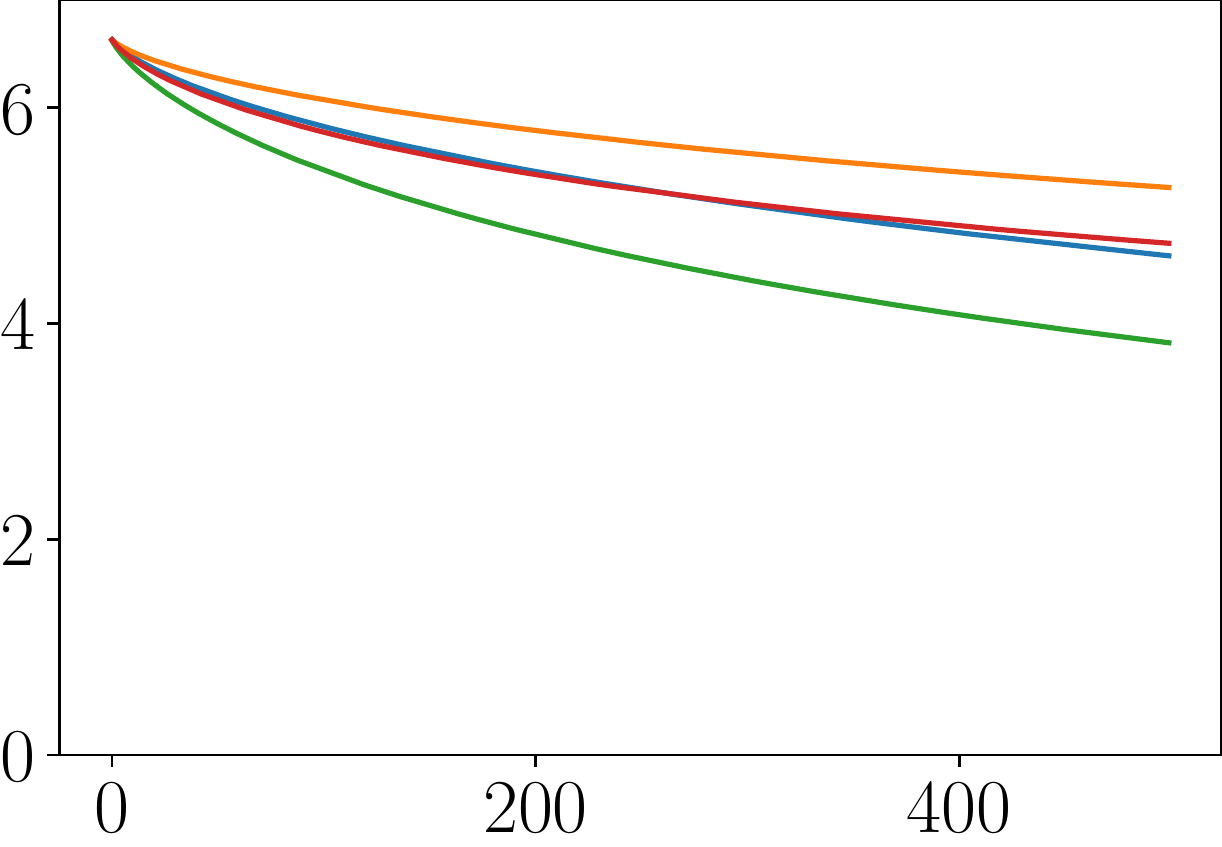} 
         \\
         & \small Epochs
         & & 
         & \small Epochs
        
    \end{tabular}
    \caption{Convergence of the direction (left) and magnitude (right) of randomly initialized weight vectors of two identical but differently initialized models for MNIST and CIFAR datasets and hidden layer widths of $9$k and $25$k.}
    \label{fig:convergence}
\end{figure}

%
\noindent%
Fig.~\ref{fig:convergence} illustrates examples of this result.
The left plot shows the trend of the cosine distance between the weight vectors of two parallel instances of a linear model, initialized with different random seeds, in terms of gradient descent epochs.
Similar to Fig.~\ref{fig:norms}, the results are reported for MNIST and CIFAR with hidden layer widths of $9$k and $25$k.
We repeat the experiment for different learning rates and different model sizes and consistently observe that the cosine distance always converges to zero with rates depending on the learning rate and model size.
This observation also holds for the two \ac{LTR} datasets that we consider in this paper (see Sec.~\ref{sec:experiments}).

\myparagraph{Start of overfitting}
In \moped linear models, the exact minimum-$\ell_2$ solution can be obtained using the Moore-Penrose inverse~\citep{prasad1992generalized}.
This exact solution corresponds to one of the zero training loss solutions of the model.
Our experiments show that, although \ac{GD} converges to the minimum-$\ell_2$ solution, it requires significantly longer training epochs than what is observed in Fig.~\ref{fig:convergence}-left.
For instance, after $100$k training epochs, the cosine distance between the \ac{GD} and minimum-$\ell_2$ solutions in different data and different model setups falls in the range of $0.1$ and $0.3$.
Comparing to Fig.~\ref{fig:convergence}, where at epoch $400$ the weights of two parallel instances have a cosine distance less than $0.01$, we see a large gap between the convergence rates of these two cases:
\begin{blockquote}
\emph{The parallel instances converge to each other long before they converge to their final value of the minimum-$\ell_2$ solution.}
\end{blockquote}
\noindent
Noticing that overfitting can start at sooner iterations than the zero training loss point, we conjecture that this early convergence coincides with the start of overfitting to the training data and provide evidence for our conjecture with a diverse set of experiments.

\myparagraph{Formal notation}
Our discussion so far can be summarized in the following formal notation.
Suppose two parallel instances of a linear network $\mathscr{L}_1$ and $\mathscr{L}_2$, with different initializations \smash{$W_1^{(0)}$} and \smash{$W_2^{(0)}$}, featurized input $A$ and target output $Y$.
Both models will converge to the minimum $\ell_2$ zero training loss solution, obtained by \smash{$W^{(\infty)}=A^+ \cdot Y$}, but, based on our empirical findings, \smash{$W_1^{(t)}$} and \smash{$W_2^{(t)}$} converge in direction at a rate considerably faster than they converge to \smash{$W^{(\infty)}$}.
Intuitively, the reason for such a behavior is that the models first learn the general structure of the training data and ignore the outlier samples~\citep{li2020gradient}.
Suppose the (unknown) set of the non-outlier (featurized) training samples that best describe the general data structure and their corresponding clean labels are shown by $\hat{A}$ and $\hat{Y}$.
Then, by learning the general structure, each model first converges to the $\hat{A}^+ \cdot \hat{Y}$ solution.
A low distant pair of \smash{$W_1^{(t)}$} and \smash{$W_2^{(t)}$} vectors indicates that the general \smash{$\hat{A}^+ \cdot \hat{Y}$} solution has been reached and, from now on, the models are converging to the zero training loss solution of $A^+ \cdot Y$.

\myparagraph{Directions or magnitudes}
Finally, to show that the above trends are only observable for the \emph{directions} and not the \emph{magnitudes}, we replace the cosine distance with the Euclidean distance in the right plot of Fig.~\ref{fig:convergence}.
As can be seen in this figure, the Euclidean distance does not converge to zero within a practical number of epochs.
More importantly, when the model overfits (which is the case in all of the shown plots), the Euclidean distance does not converge to a fixed value across different datasets, model sizes and learning rates, making it difficult to base an early stopping criterion on it.\looseness=-1

\subsection{\acl{CDC}}
\label{subsec:ourmethod}
Based on what we have discussed in the previous section, we can state our simple rule for early stopping:

\begin{blockquote}
    \emph{Train in parallel two identical instances of a linear model, initialized with different random seeds, and stop training when the L-curve obtained from the cosine distance between their weight vectors in terms of training epochs passed its corner.}
\end{blockquote}

\noindent
Findings of~\citet{li2020gradient} (as discussed in Sec.~\ref{sec:related}) support this idea to some extent.
They show that in the presence of corrupted labels, the first iterations of gradient descent ignores them and tries to fit on the correct labels.
During this phase, the weights remain close to their initial values.
But as the iterations proceed, the model starts to overfit the corrupted labels and the weights move far from their initial value.
Here, our empirical observation suggests a more specific behavior: networks converge to directions far from their initialization, but this new direction is the same for identical but differently initialized networks and it is not necessarily the direction corresponding to zero training loss.

\begin{algorithm}[b]
    \caption{\ourmethod}
    \label{alg:cdrule}
    \SetAlgoLined
    \KwIn{$D$, $\eta$, $T_{\max}$}
    \KwOut{Early Stopping Epoch}
    Build $\mathscr{N}_1$ and $\mathscr{N}_2$ with weights $W_1$ and $W_2$ of length $D$\\
    Initialize $W_1$ and $W_2$ with different random seeds\\
    \For{$t = 0$ to $ T_{\max}$}{
        Update $W_1$ and $W_2$ using first order methods with learning rate $\eta$\\
        $\delta^{(t)}=\cos(W_1^{(t)}, W_2^{(t)})$\\ 
        \If{$\delta^{(t)}$ is a L-curve corner}{
            Return $t$
        }
    }
    Return $T_{\max}$
\end{algorithm}

\myparagraph{Using parallel instances}
In order to materialize this idea for early stopping, we need a robust method to detect when the weights of the linear model are \emph{converged} far enough from their initial value, and the model is starting to overfit.
Two extreme cases are clear:
\begin{enumerate*}[label=(\roman*)]
    \item at the beginning iterations, the weights are close to the initialization and the model underfits; but
    \item after too many iterations, the weights are far from the initialization and the model overfits.
\end{enumerate*}
Knowing these two extreme cases is not enough.
Instead, we are interested in finding the transition point between them.
Looking at one model and measuring the distance of its weights to their initial value is not sufficient for detecting the transition point, as it is not clear \emph{how far} from the initialization point the transition occurs.
Our experiments reveal that this issue is solvable by comparing two identical but differently initialized models, instead of only looking at a single model.
Two identical models, starting from two different initialization points, converge to the same zero training loss point.
As a consequence of this shared final destination, they are converging to each other too.
The closer they get to each other, they will be closer to their shared final destination of zero training loss, and they overfit more to the train data.
Therefore, the similarity of these models at each iteration can be used as a proxy for detecting their degree of overfitting.
Starting from a cosine distance close to one (as two random high dimensional vectors are almost surely orthogonal), the two models converge to each other as they learn the training data.
After the general structure of the training data has been learned, the two models start to overfit.
Our empirical observations on four datasets with different modalities and models with different sizes and learning rates suggest that the changing rate of the cosine distance differs before and after the sweet point of maximum generalization:
the cosine distance decreases sharply before that point, but slowly after it.
That is why we observe an L-curve similar to what was shown in Fig.~\ref{fig:lcurve}.
We take the corner of such an L-curve as the maximum generalization point.

\myparagraph{The algorithm}
Algorithm~\ref{alg:cdrule} shows the formal pseudo-code for \ourmethod.
The algorithm gets as input the number of parameters ($D$), learning rate ($\eta$) and the maximum number of epochs ($T_{\max}$).
Despite its simplicity, our extensive experiments show the effectiveness of \ourmethod compared to existing methods for a wide range of learning rates.
For line 6 of Algorithm~\ref{alg:cdrule}, we test two simple methods: 
\begin{enumerate*}[label=(\roman*)]
    \item the maximum curvature condition; and
    \item a fixed threshold for all cases.
\end{enumerate*}
Next, we discuss and compare these two methods for corner detection in more detail.

\subsection{Corner Detection}
\label{sec:method:corner}
As discussed in Sec.~\ref{sec:related}, the L-curve corner detection is a popular regularization method for solving systems of linear equations~\citep{hansen1998rank, castellanos2002triangle, kim2007regularization}.
Here, we show our method has a low sensitivity to the hyper parameters of corner detection methods and in all of our tested datasets and model setups, it is safe to consider a fixed threshold for the cosine distance as the early stopping point.

\myparagraph{Maximum curvature}
Let $c_i$ be the cosine distance between the two models at iteration $i$.
Since we are working with a discrete series, we need a step size $\Delta$ to act as the sampling points in our curves: each $c_k$ value corresponds to the virtual $(i\cdot\Delta, c_i)$ point in the 2-D plain.
Using this terminology, curvature is defined as follows (adapting~\citep{hansen1998rank} to our terminology):
\begin{equation}
    \label{eq:curvatureformula}
    \kappa_i = \Delta \cdot \frac{c_i - c_{i-2}}{\left(\Delta^2 + \left(c_i - c_{i-1}\right)^2\right)^{\frac{3}{2}}}.
\end{equation}
\noindent
Here, $\Delta$ is a hyper parameter, modeling the compression of the x-axis in plots similar to Fig.~\ref{fig:lcurve}.
In practice, the $\left\{c_i\right\}$ series may not be smooth.
So, before computing Eq.~\eqref{eq:curvatureformula}, we first apply Gaussian kernel smoothing to the series.
The iteration with maximum $\kappa_i$ is returned as the corner point in line $6$ of Algorithm~\ref{alg:cdrule}.
For the hyper parameter $\Delta$, our experiments show a very low sensitivity as will be discussed next.
Intuitively, the discretization step should be in the order of the learning rate. 
In Fig.~\ref{fig:maxcurve_sensitivity} we show two examples on MNIST (left) and CIFAR (right) datasets.
The solid vertical black line is the maximum curvature obtained by setting $\Delta$ equal to the learning rate ($0.004$ and $0.002$ for MNIST and CIFAR, respectively).
The shaded area contains the epochs with maximum curvature for different $\Delta$ values, ranging from $\frac{lr}{2}$ to $5\times lr$.
The test performance drops, for this wide range of hyper parameter, are only 97\% and 93\% for MNIST and CIFAR, respectively.
Various experiments on other datasets (see Sec.~\ref{sec:experiments}) and with different model sizes and learning rates lead to similar results.

\myparagraph{Fixed threshold}
We use the cosine distance to measure the similarity of two identical but differently initialized instances of a model and want to decide about the starting point of overfitting from this similarity.
Therefore, it is natural to set a fixed threshold for this similarity and early stop when the instances \emph{intersect}, i.e. become more similar than that threshold.
Here, it is not theoretically clear what choice for the fixed threshold leads to the optimal results and whether this threshold should depend on the dataset and the model setup or not.
However, our empirical observations on four datasets from two different domains (i.e. vision and LTR) and on a wide range of model sizes and learning rates show that a threshold of $\theta=0.2$ works fine in most of the cases.
This threshold value also works for our multi-layer experiments (to be discussed later in Sec.~\ref{sec:multilayer}).
We leave further analysis of why a threshold of $\theta=0.2$ for the cosine distance works fine for a very wide range of datasets and model setups to future work.
Now, similar to what we have shown previously in the maximum curvature method, we show that our method has a low sensitivity to this $\theta$ hyper parameter.
In Fig.~\ref{fig:threshold_sensitivity} we show two examples on MNIST (left) and CIFAR (right) datasets.
The solid vertical black line is where the cosine distance crosses $\theta=0.2$.
The shaded area contains the epochs with $0.1\leq\theta\leq0.5$.
The test performance drops, for this wide range of hyper parameter, are only 98\% and 94\% for MNIST and CIFAR, respectively.
Various experiments on other datasets (see Sec.~\ref{sec:experiments}) and with different model sizes and learning rate lead to similar results.

As the fixed threshold method has a lower sensitivity and it is simpler, we only report the performance of our criterion with this corner detection method and set $\theta=0.2$.
We should also stress that the maximum curvature method with $\Delta=lr$ leads to similar results.

{
    \setlength{\tabcolsep}{0.1em}
\begin{figure}[t]
    \centering
    \begin{tabular}[]{lccr}

        \rotatebox[origin=lt]{90}{\hspace{1.2em} \small Accuracy}
        &
        \includegraphics[scale=0.25]{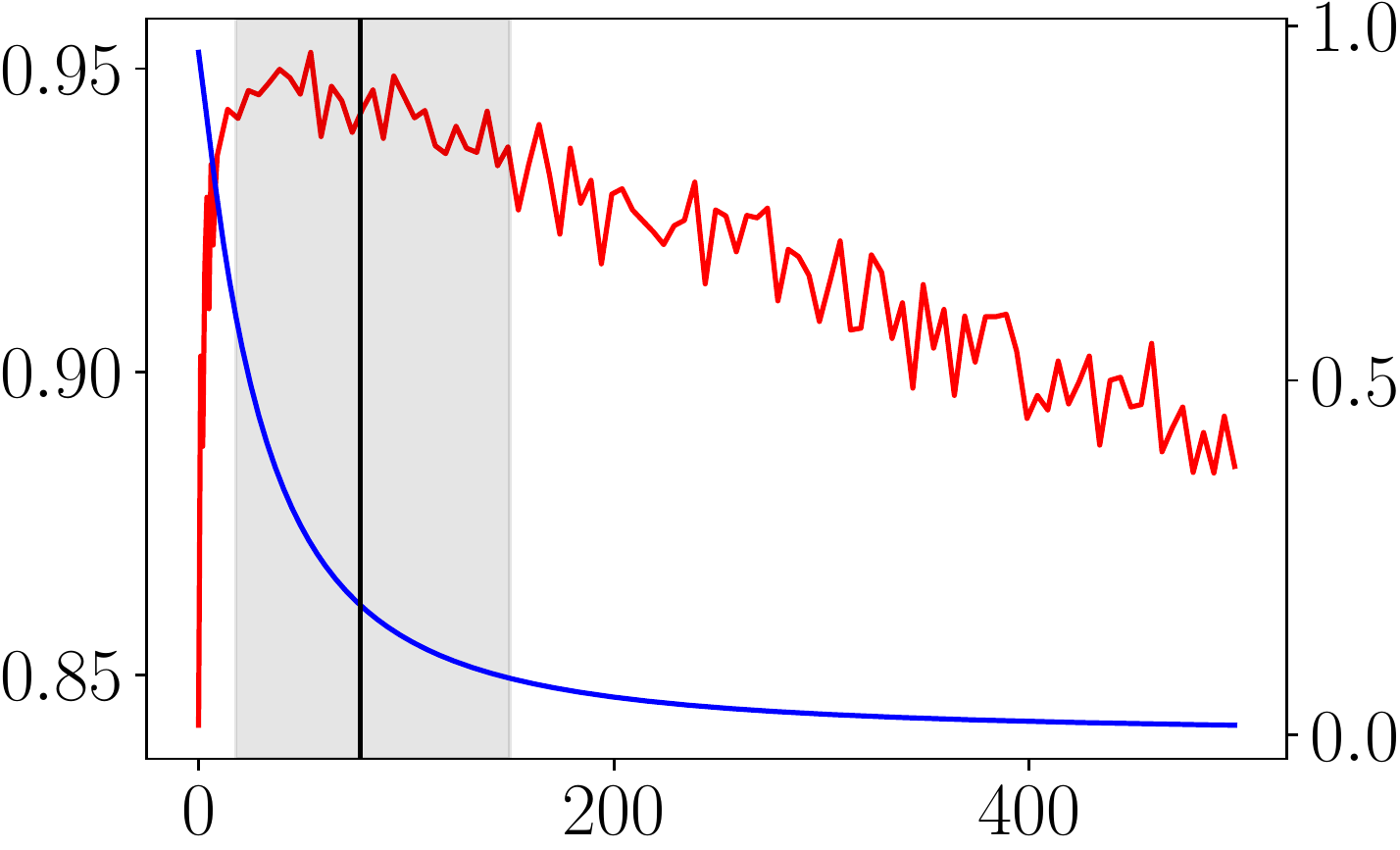}
        &
        \includegraphics[scale=0.25]{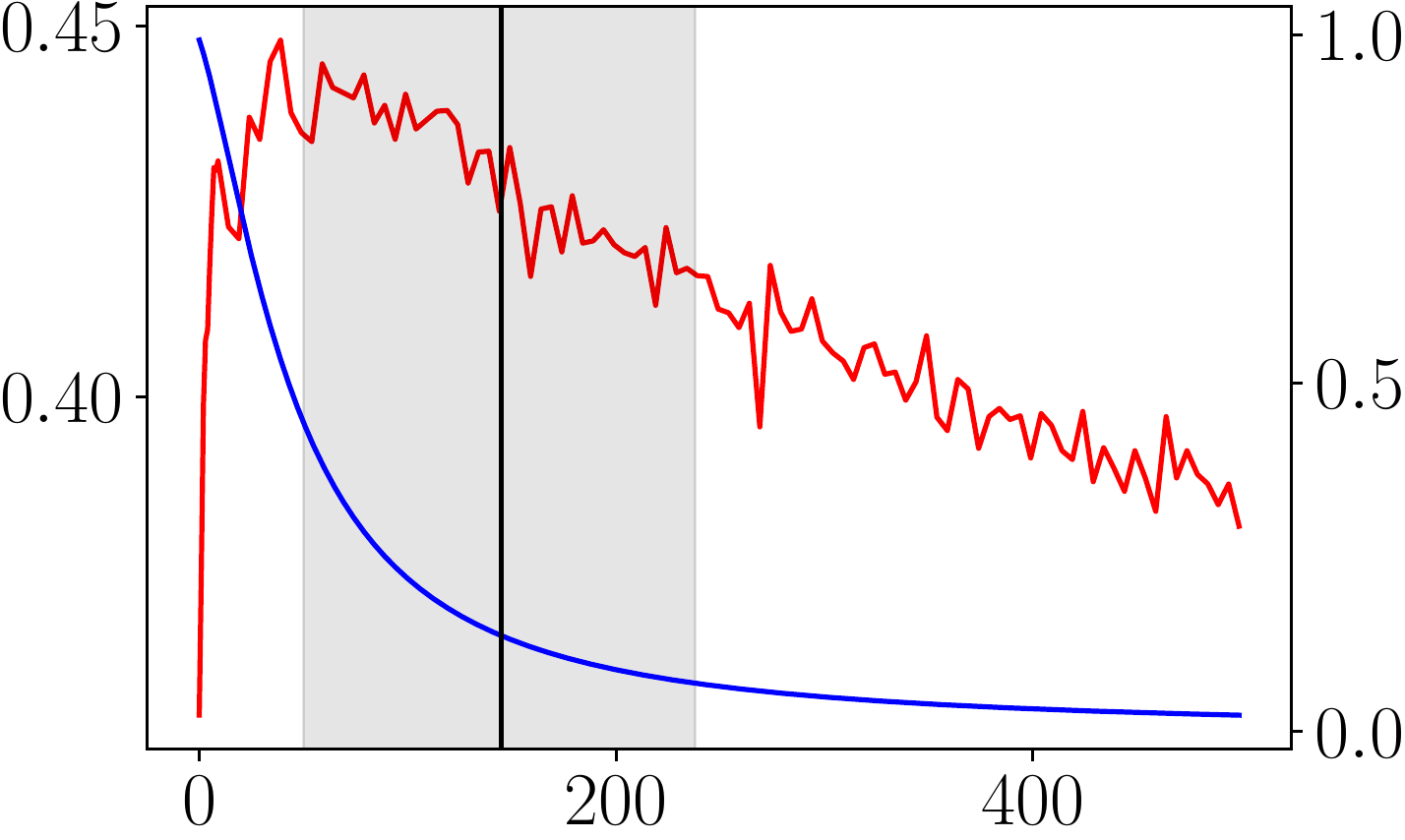}
        &
        \rotatebox[origin=lt]{90}{\hspace{0.4em} \small Cosine distance}  \\
        & \small Epochs
        & \small Epochs \\
        &
        \multicolumn{2}{c}{\includegraphics[scale=0.40]{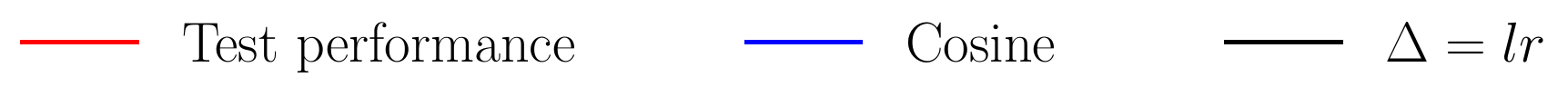}}
        
    \end{tabular}
    \caption{Sensitivity of the maximum curvature corner detection method to $\Delta$ on MNIST (left) and CIFAR (right) datasets with 50\% random labels.}
    \label{fig:maxcurve_sensitivity}
\end{figure}
}
{
    \setlength{\tabcolsep}{0.1em}
\begin{figure}[t]
    \centering
    \begin{tabular}[]{lccr}

        \rotatebox[origin=lt]{90}{\hspace{1.2em} \small Accuracy}
        &
        \includegraphics[scale=0.25]{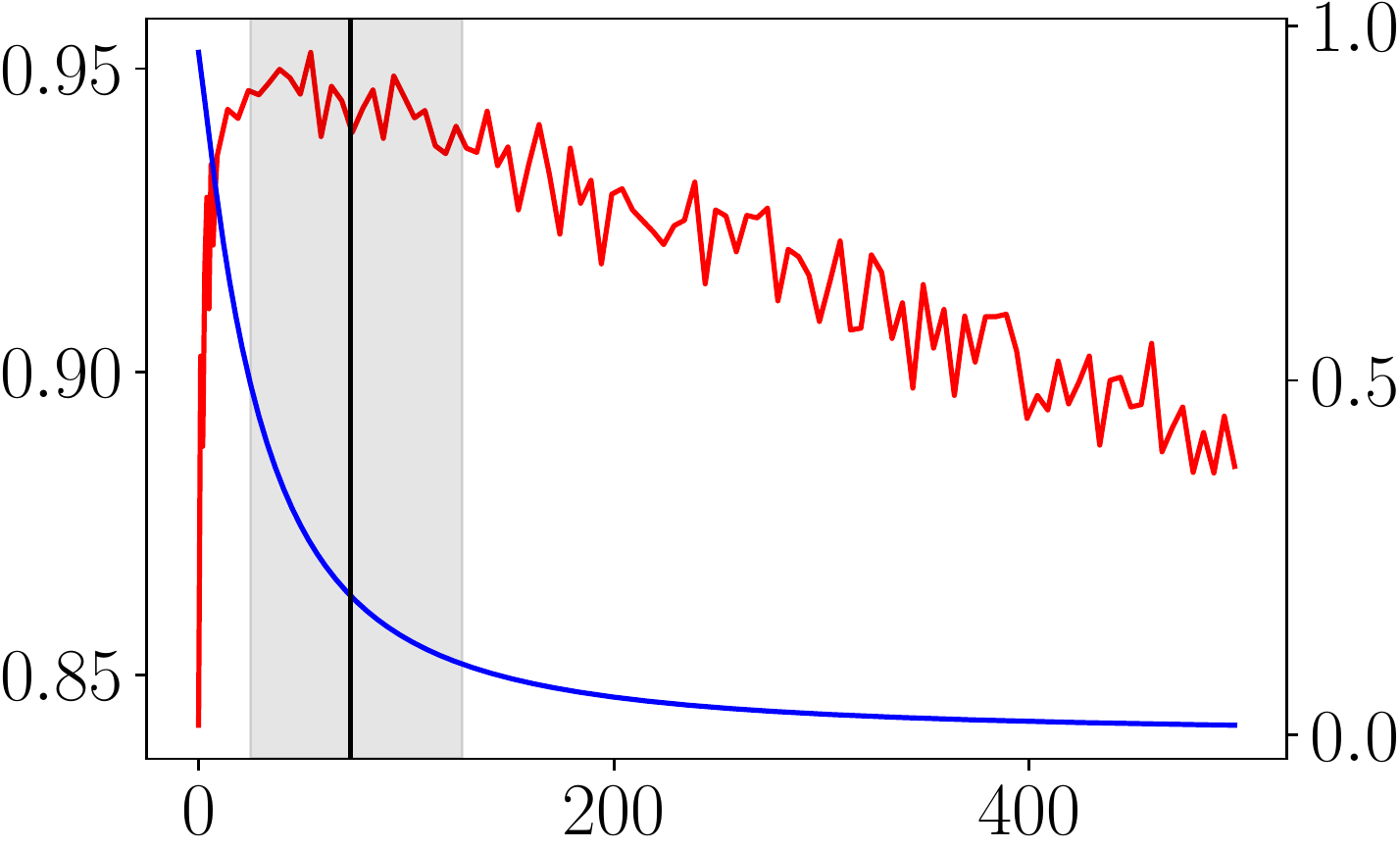}
        &
        \includegraphics[scale=0.25]{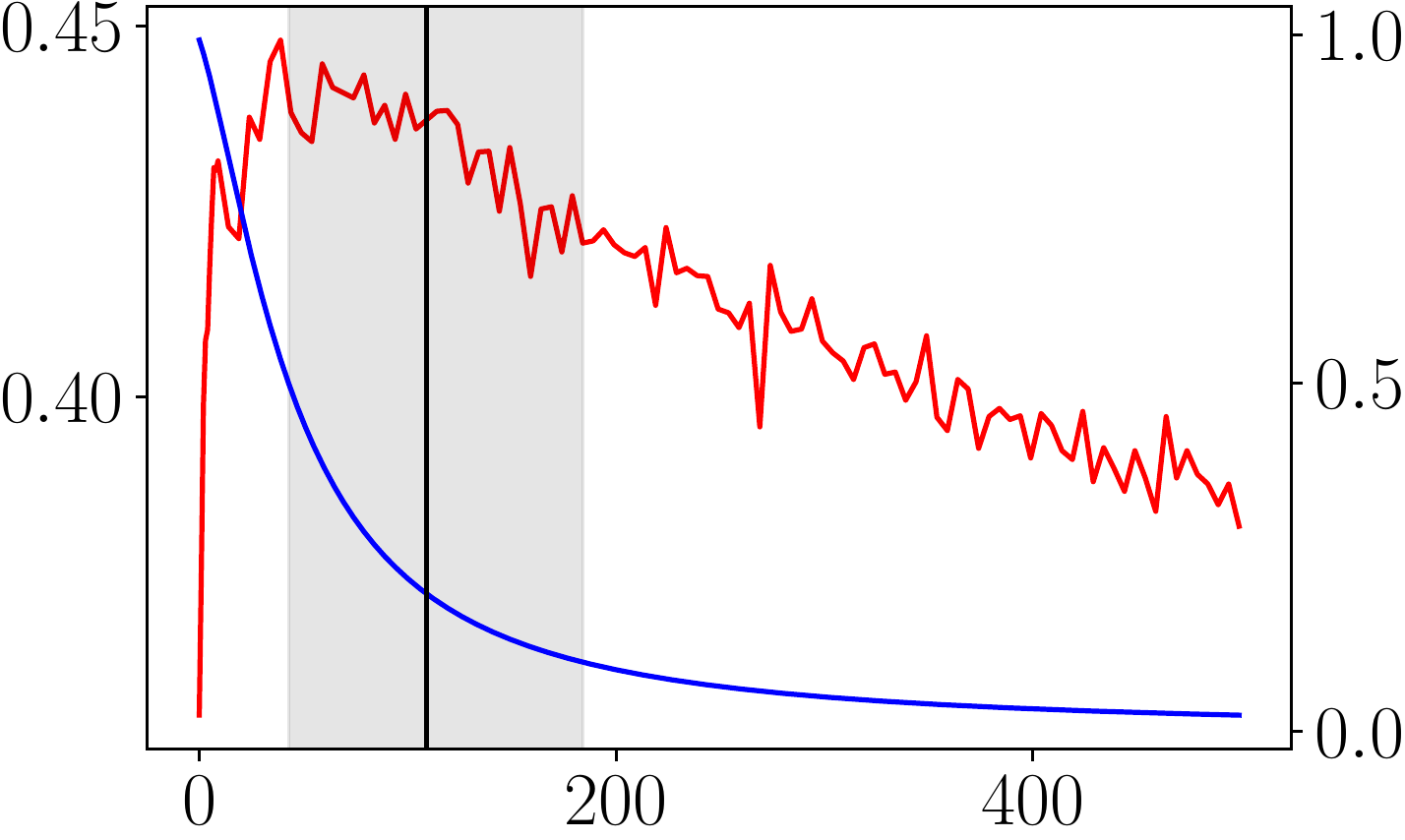}
        &
        \rotatebox[origin=lt]{90}{\hspace{0.4em} \small Cosine distance}  \\
        & \small Epochs
        & \small Epochs \\
        &
        \multicolumn{2}{c}{\includegraphics[scale=0.40]{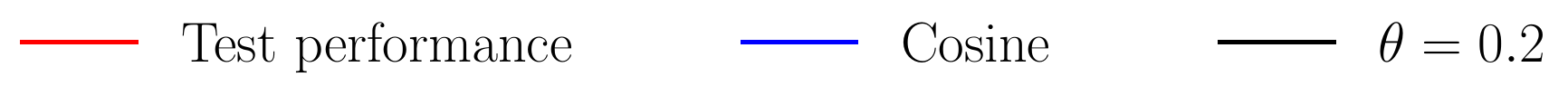}}
        
    \end{tabular}
    \caption{Sensitivity of fixed threshold corner detection method to $\theta$ on MNIST (left) and CIFAR (right) datasets with 50\% random labels.}
    \label{fig:threshold_sensitivity}
\end{figure}
}

\subsection{Multi-Layer Networks}
\label{sec:multilayer}

As discussed in Sec.~\ref{sec:multilayer:notations}, in a multi-layer network, the pre-last layer activation matrix depends on trainable parameters.
As such, for two parallel instances $\mathscr{N}_1$ and $\mathscr{N}_2$ with different initializations, we have different, time-dependent activation matrices \smash{$A_1^{(t)}$} and \smash{$A_2^{(t)}$}.
The target vector $Y$, on the other hand, is shared for both instances.
This means that \smash{$W_1^{(t)}$} and \smash{$W_2^{(t)}$} should map different inputs \smash{$A_1^{(t)}\neq A_2^{(t)}$} to a shared output.
Consequently, comparing \smash{$W_1^{(t)}$} and \smash{$W_2^{(t)}$} is comparing apples and oranges.
Between the two instances, only the predictions \smash{$\Tilde{Y}_1^{(t)}$} and \smash{$\Tilde{Y}_2^{(t)}$} are comparable, since they are both produced to match the same target label vector $Y$.
In order to settle this issue, we try to answer the following counterfactual question:
\begin{blockquote}
\emph{Had $\mathscr{N}_2$ been given the activation matrix of $\mathscr{N}_1$, that is, \smash{$A_1^{(t)}$} instead of \smash{$A_2^{(t)}$}, what would have been its weight vector to produce the prediction \smash{$\Tilde{Y}_2^{(t)}$}?}
\end{blockquote}
{
    \setlength{\tabcolsep}{0.7em}
\begin{figure}[t]
    \centering
    \begin{tabular}[]{lcc}

        \rotatebox[origin=lt]{90}{\hspace{0.2em} \small Cosine distance}
        &
        \includegraphics[scale=0.25]{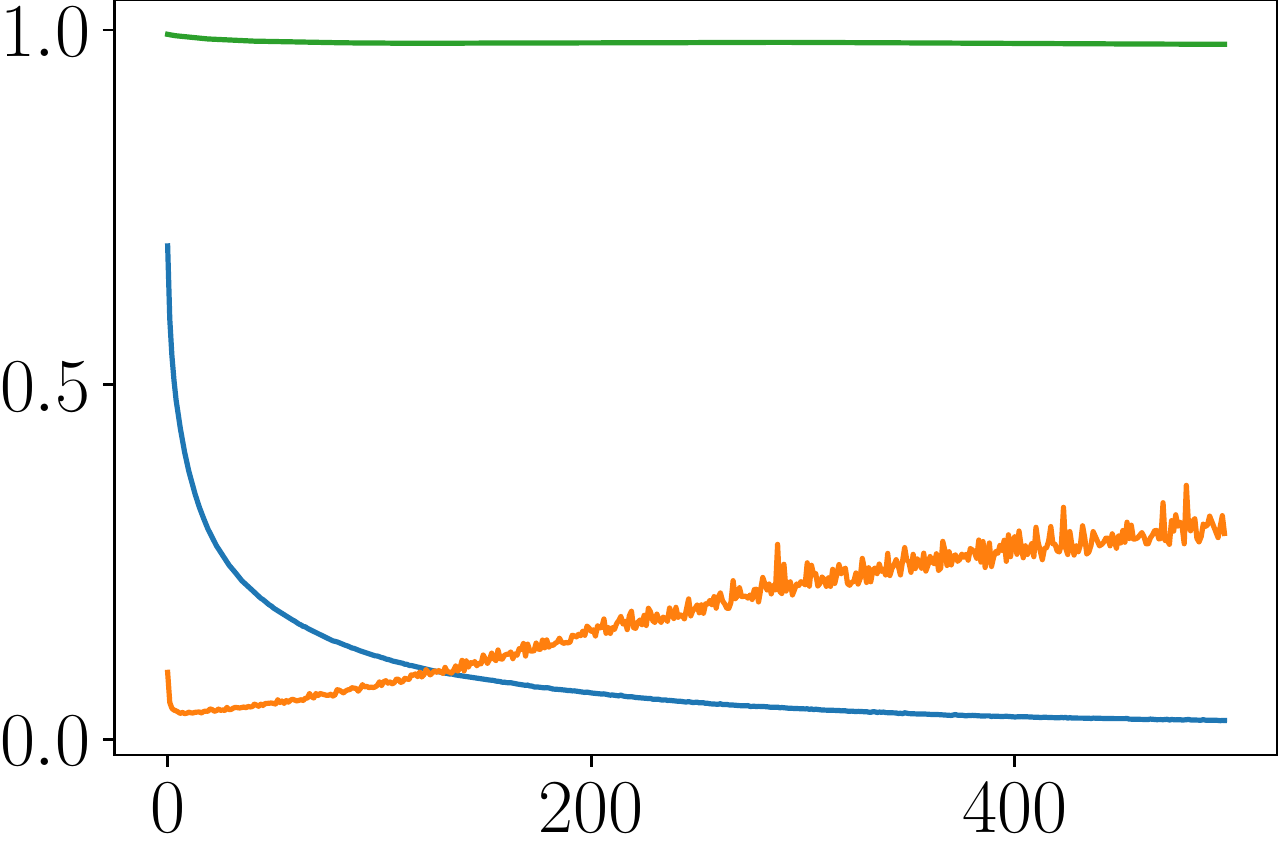}
        &
        \includegraphics[scale=0.25]{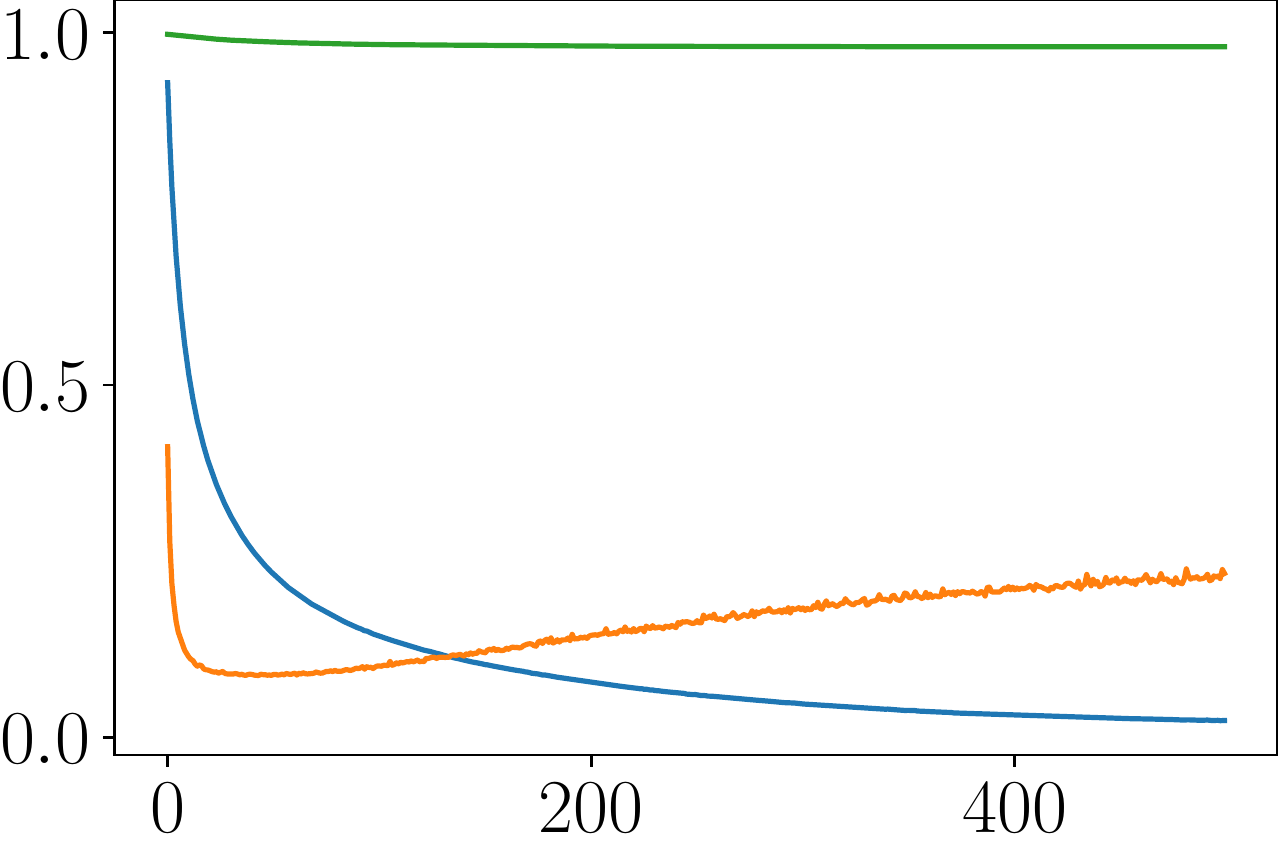}  \\
        & \small Epochs
        & \small Epochs \\
        &
        \multicolumn{2}{c}{\includegraphics[scale=0.40]{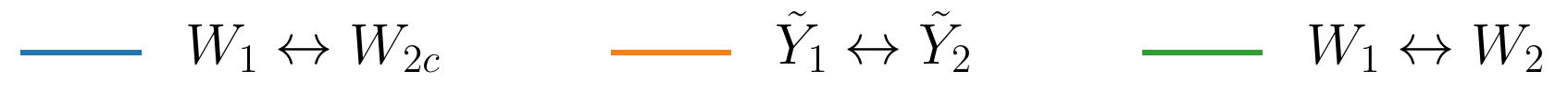}}
        
    \end{tabular}
    \caption{Cosine distance of different vectors for 2-layer networks on MNIST (left) and CIFAR (right) datasets with 50\% random labels.}
    \label{fig:output_distance}
\end{figure}
}
\noindent
We call \smash{$W_{2c}^{(t)}$} the \emph{counterfactual weight} of $\mathscr{N}_2$ as the answer to the above question:
\begin{equation}
\label{eq:counterfactual}
    A_1^{(t)} \cdot W_{2c}^{(t)} = \Tilde{Y}_2^{(t)}.
\end{equation}
\noindent
The best solution in terms of $\ell_2$ error to Eq.~\eqref{eq:counterfactual} is given by the Moore-Penrose inverse:
\begin{equation}
\label{eq:counterfactualsolution}
     W_{2c}^{(t)} = A_1^{(t)+} \cdot \Tilde{Y}_2^{(t)}.
\end{equation}
\noindent
It is worth noting that \smash{$W_{1}^{(t)} = A_1^{(t)+} \cdot \Tilde{Y}_1^{(t)}$}.
Or in words, \smash{$W_{1}^{(t)}$} and \smash{$W_{2c}^{(t)}$} are projections of predictions \smash{$\Tilde{Y}_1^{(t)}$} and \smash{$\Tilde{Y}_2^{(t)}$}, using the same matrix of $A_1^{(t)+}$.

In Sec.~\ref{subsec:convergence} we argued that the networks first learn the general structure of data by fitting to the ``set of the non-outlier samples that best describe the general data structure and their corresponding clean labels, shown by $\hat{A}$ and $\hat{Y}$''.
Here, in case of multi-layer networks, when $\mathscr{N}_1$ and $\mathscr{N}_2$ are fit to $\hat{Y}$, it means that \smash{$\Tilde{Y}_1^{(t)}$} and \smash{$\Tilde{Y}_2^{(t)}$} agree on the (unknown) $\hat{Y}$ part of their predictions.
But, since they do not overfit yet, they have random predictions about the outliers.
Consequently, measuring the distance between \smash{$\Tilde{Y}_1^{(t)}$} and \smash{$\Tilde{Y}_2^{(t)}$} is simply not informative.
However, when they are projected using \smash{$A_1^{(t)+}$}, they should be mapped to weight vectors close to \smash{$\hat{A}^+ \cdot \hat{Y}$}.

We experimentally confirm the above theory by showing that the cosine distance between \smash{$\Tilde{Y}_1^{(t)}$} and \smash{$\Tilde{Y}_2^{(t)}$} diverges instead of converging, while their projections \smash{$W_{1}^{(t)}$} and \smash{$W_{2c}^{(t)}$} converge.
Fig.~\ref{fig:output_distance} shows two examples of the results of our experiments on MNIST (left) and CIFAR (right) datasets.
All the other setups for a 2-layer network lead to similar observations.
Here, we plot the cosine distance between three pairs of vectors:
\begin{enumerate*}[label=(\roman*)]
    \item Weight vectors \smash{$W_{1}^{(t)}$} and \smash{$W_{2}^{(t)}$} are not comparable and their direction remains orthogonal during training.
    \item Predicted vectors \smash{$\Tilde{Y}_1^{(t)}$} and \smash{$\Tilde{Y}_2^{(t)}$} do not converge in direction, because $\mathscr{N}_1$ and $\mathscr{N}_2$ first try to fit to $\hat{Y}$ and predict differently for the other (noisy or outlier) part of the samples.
    \item Finally, the projected vectors \smash{$W_{1}^{(t)}$} and \smash{$W_{2c}^{(t)}$} are comparable and converge in direction, similar to what we observe in linear models (e.g., Fig.~\ref{fig:lcurve}).
\end{enumerate*}

The limitation of this method is when \smash{$A_1^{(t)}$} is rank deficient, so the solution given by Eq.~\eqref{eq:counterfactualsolution} is not correct.
In our experiments, we encountered such situations for mildly \moped networks with four or more layers.
We leave further investigations of when this happens and what can be done to fix it for future work.

\section{Experiments}
\label{sec:experiments} 
In this section we explain the datasets and model used in our experiments.
We also explain the runs that we use to compare and show the effectiveness of our~\ourmethod rule in generalization.

\vspace{-0.3em}
\subsection{Setup}
\label{subsec:setup} 
\textbf{\em Image datasets.}
We use two image classification benchmark data-sets: MNIST and CIFAR-10.
Both datasets consist of 10 classes.
MNIST contains $60$k training and $10$k test grayscale $28\times 28$ images of handwritten digits, while CIFAR-10 has $50$k training and $10$k test RGB $32\times 32$ object images.
Following~\citep{forouzesh2021disparity}, we make these datasets noisy by changing 50\% of the labels to random values.

\myparagraph{LTR datasets}
As a regular choice in \ac{LTR} research~\citep{joachims2017unbiased, jagerman2019comparison, vardasbi2020when},
we use two popular \ac{LTR} datasets: Yahoo! Webscope~\citep{Chapelle2011} and MSLR-WEB30k~\citep{qin2013introducing}.
For each query, these datasets contain a list of documents with human-generated 5-level relevance labels.
After cleaning, the query-document feature vectors of Yahoo!\ and MSLR are 501 and 131 in length, respectively.
For our experiments we randomly select 100 and 30 training queries from Yahoo! and MSLR to have around\footnote{The exact number is different for different random selections.} 3.3k training samples (i.e., document-query pairs) for each dataset.
We repeated experiments with smaller selections (down to 330 training documents) and got similar results.
For both datasets, we use the test set split as provided in the original data.
The Yahoo! dataset contains 6.7k test queries and 163k test documents; MSLR contains 6.1k and 749k query and documents in its test set.\looseness=-1

\myparagraph{Model}
For image classification tasks, we report results from \moped models with three different sizes:
\begin{enumerate*}[label=(\roman*)]
    \item {\bfseries Small};
     \item {\bfseries Medium}; and
      \item {\bfseries Large}.
\end{enumerate*}
For linear models, we respectively use $9k$, $25k$, and $50k$ featurization, leading to $90k$, $250k$, and $500k$ parameters.
For two-layer networks, we respectively use $150$, $350$, and $700$ hidden layer widths, leading to $119k$, $278k$, and $556k$ parameters for MNIST and $462k$, $1M$, and $2M$ parameters for CIFAR-10 datasets.
The last layer of image classification tasks has a dimension of $D \times 10$, because of the 10 classes.
For our method, we work with \emph{vectors}.
As a work-around, we consider the $D \times 10$ matrix as $10$ vectors of length $D$ and take the average cosine distance between the $10$ vector pairs of parallel instances.
We use the cross-entropy loss for image classification.

For LTR tasks, we only report the results with $10k$ parameters due to space limitations.
To show the effectiveness of our method with different loss functions, we report the results with two popular LTR loss functions, namely the pointwise {RMSE} loss and the listwise ListNet loss~\citep{cao2007learning}.

\myparagraph{Metric}
We report the classification accuracy on the test set for image classification tasks.
For \ac{LTR} methods, we evaluate models in terms of their NDCG@10 performance on the test set.


\subsection{Baselines}
\label{subsec:baselines}
We compare our \ourmethod method with the following existing work:
\begin{enumerate}[leftmargin=*,nosep]
    \item {\bfseries {Cross Validation (CV):}} The traditional way for early stopping.
    For this baseline, we use 5-fold cross validation to approximate the generalization drop: if for five consecutive epochs the cross validation performance did not improve, we stop the training.
    \item {\bfseries Evidence-Based (EB):} The method proposed in~\citep{mahsereci2017early} that uses the gradient size to estimate the transition point in the generalization curve.
    \item {\bfseries Gradient Disparity (GD):} A recent method proposed in~\citep{forouzesh2021disparity} based on gradient disparity.
    If the gradient disparity is increased for five consecutive epochs, the training is stopped.
    \item {\bfseries Oracle:} The skyline performance obtained by monitoring the test performance for 500 epochs and selecting the epoch with maximum test performance.
    We choose the maximum of 500 epochs because in the settings of our experiments the maximum test performance occurs well before 500 epochs with probability almost equal to one.
    Note that this is not a realistic run as we do not have access to test labels in reality.
    We include this to compare the gap between different runs and the ideal case.
\end{enumerate}


\begin{figure*}[t]
    \centering
    \begin{tabular}[]{cc}
    \begin{tabular}[]{lcccc}
        & \small Small & \small Medium & \small Large \\
        \rotatebox[origin=lt]{90}{\hspace{2em} \small Accuracy}
        &
        \includegraphics[scale=0.33]{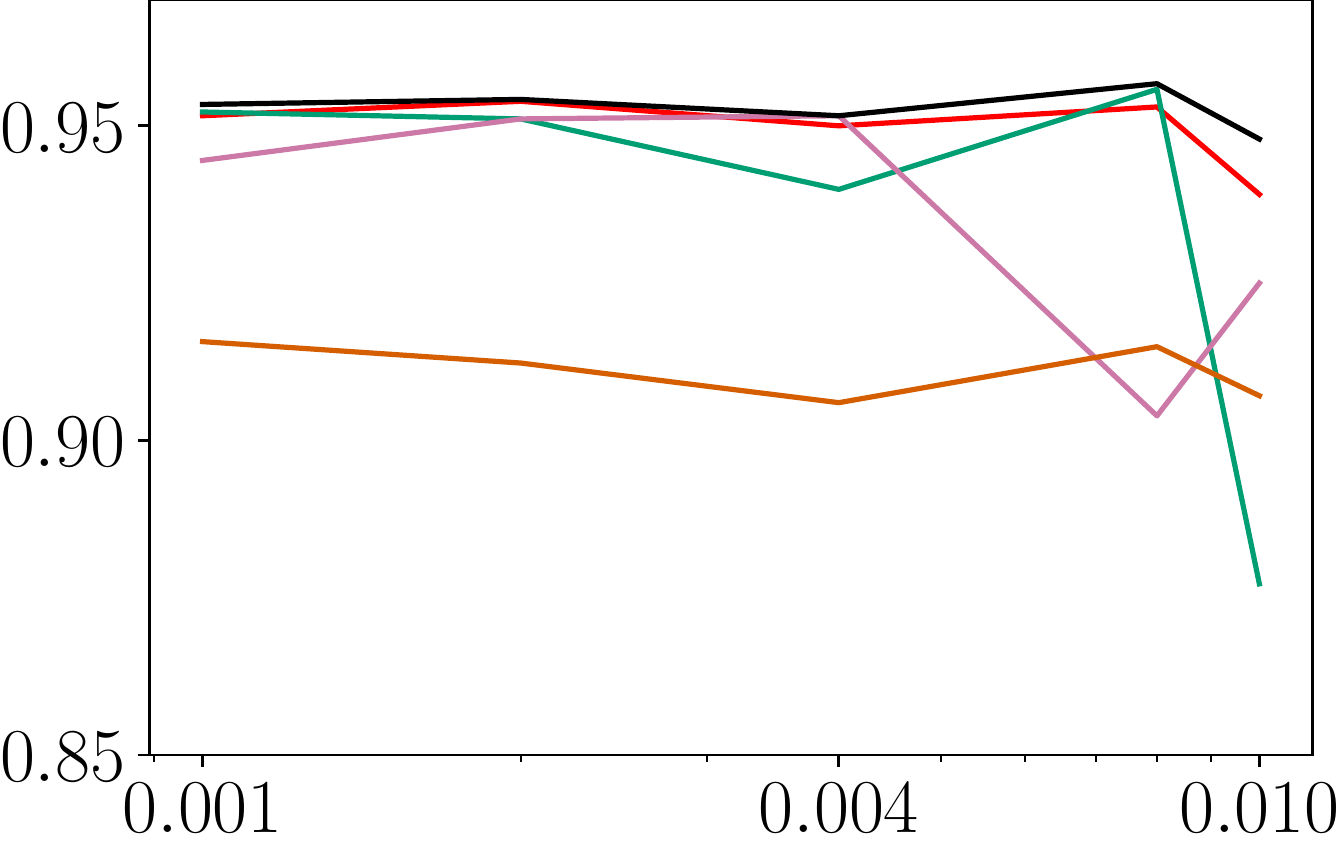}
        &
        \includegraphics[scale=0.33]{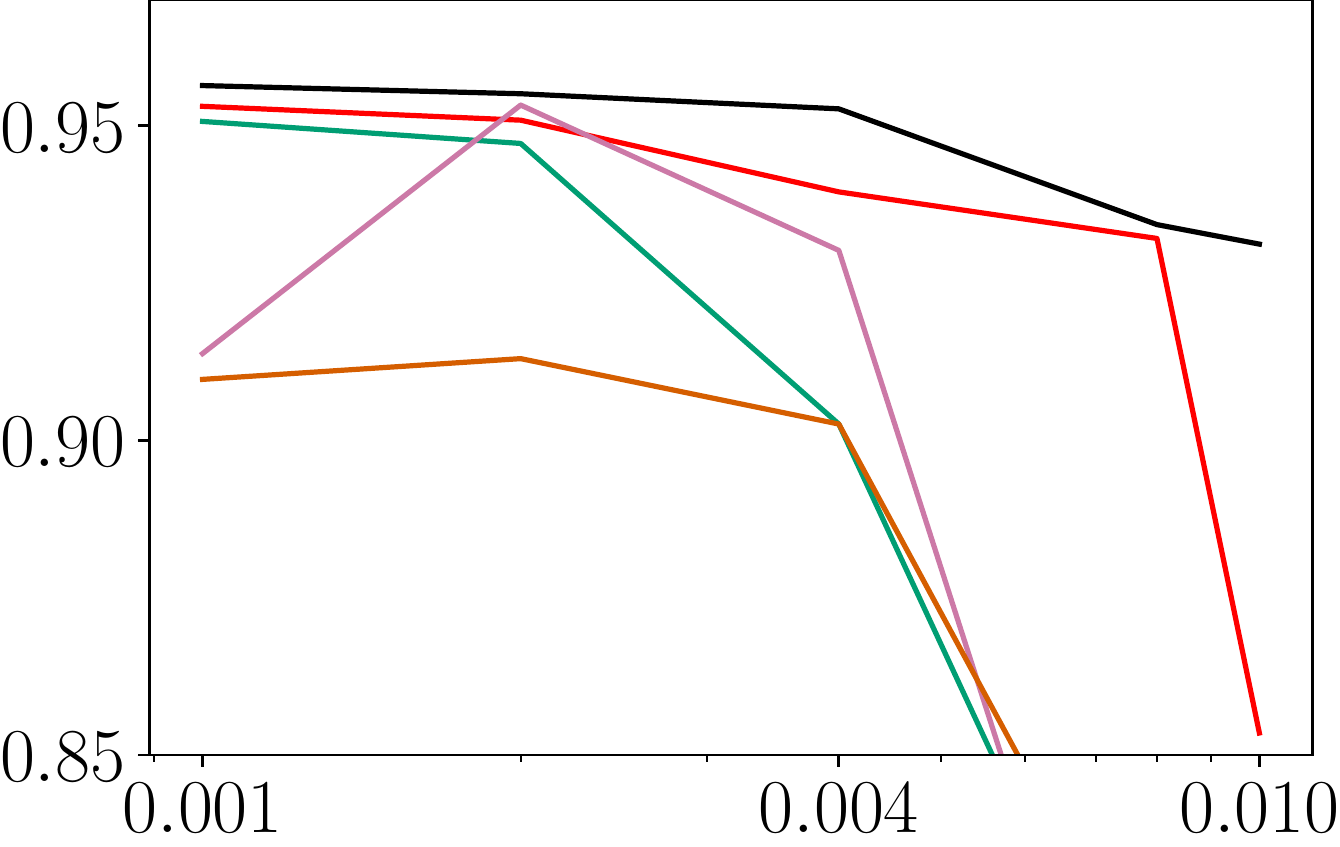}
        &
        \includegraphics[scale=0.33]{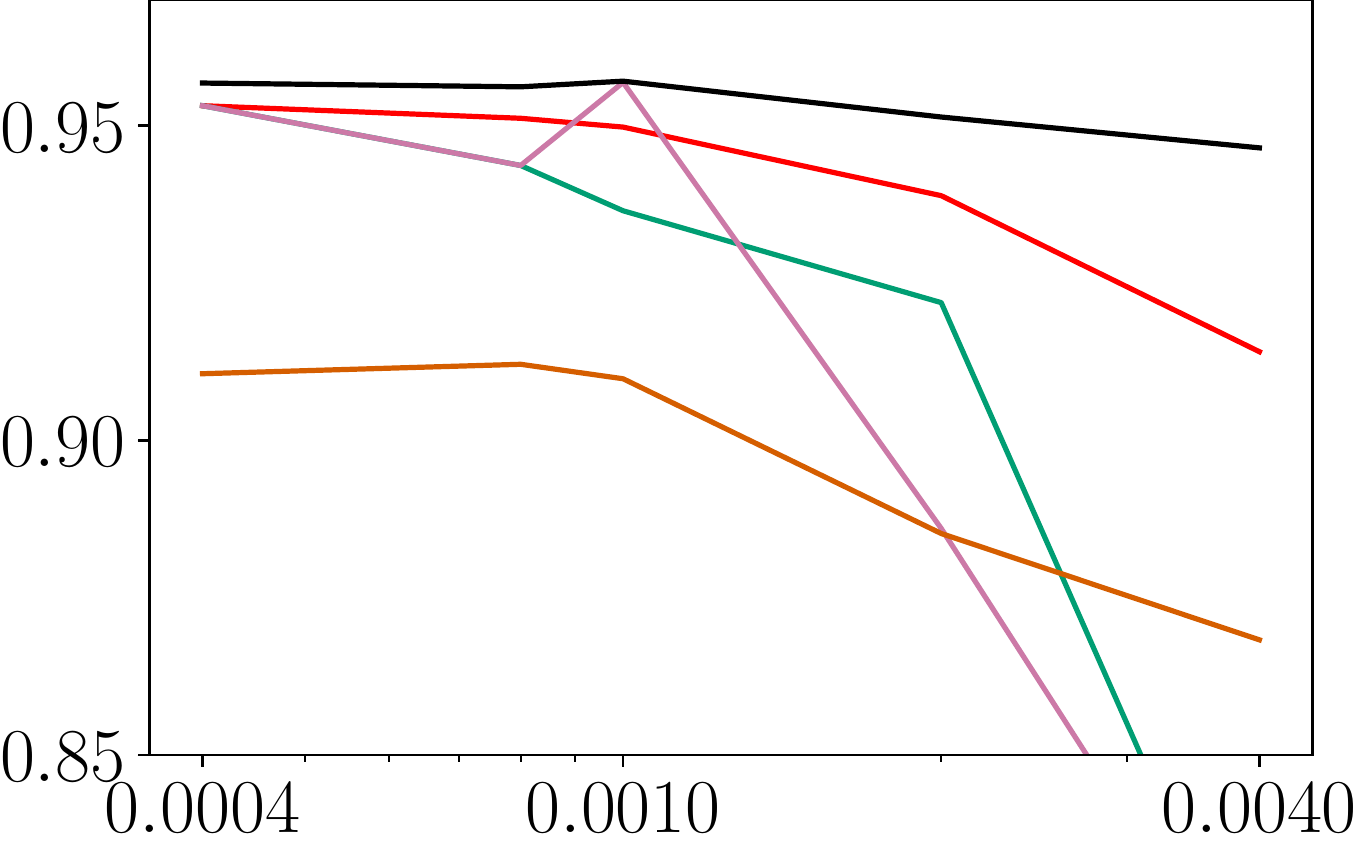}
        &
        \\
        \rotatebox[origin=lt]{90}{\hspace{2em} \small Accuracy}
        &
        \includegraphics[scale=0.33]{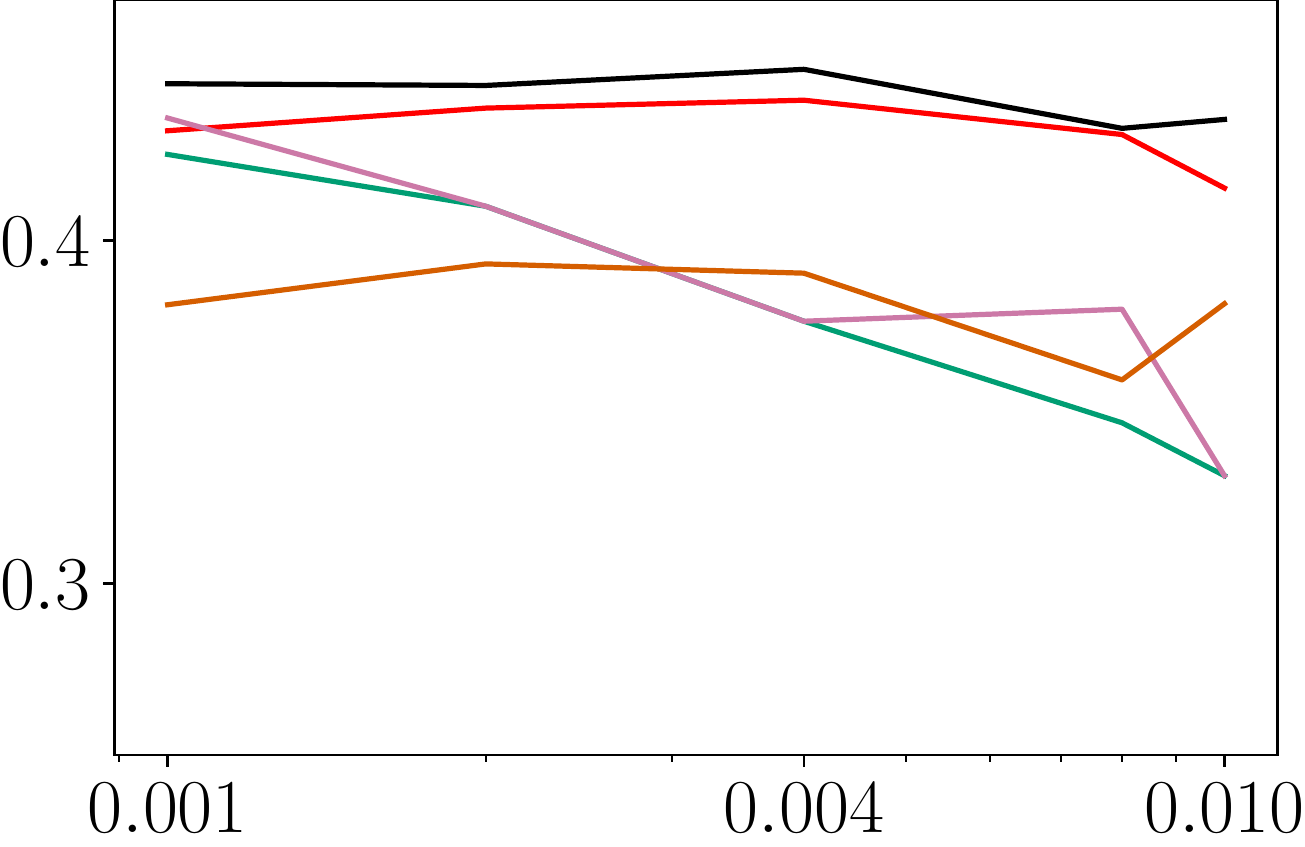}
        &
        \includegraphics[scale=0.33]{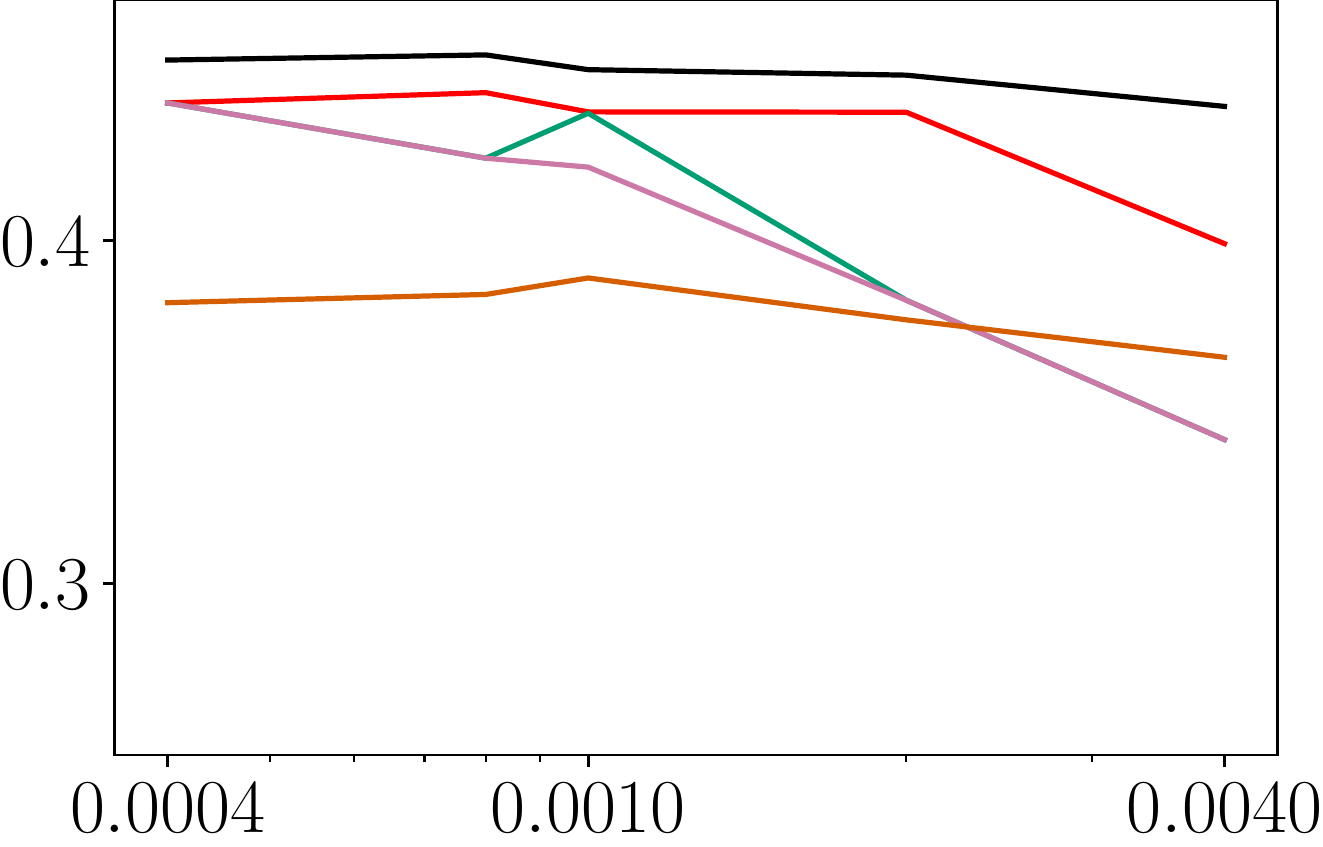}
        &
        \includegraphics[scale=0.33]{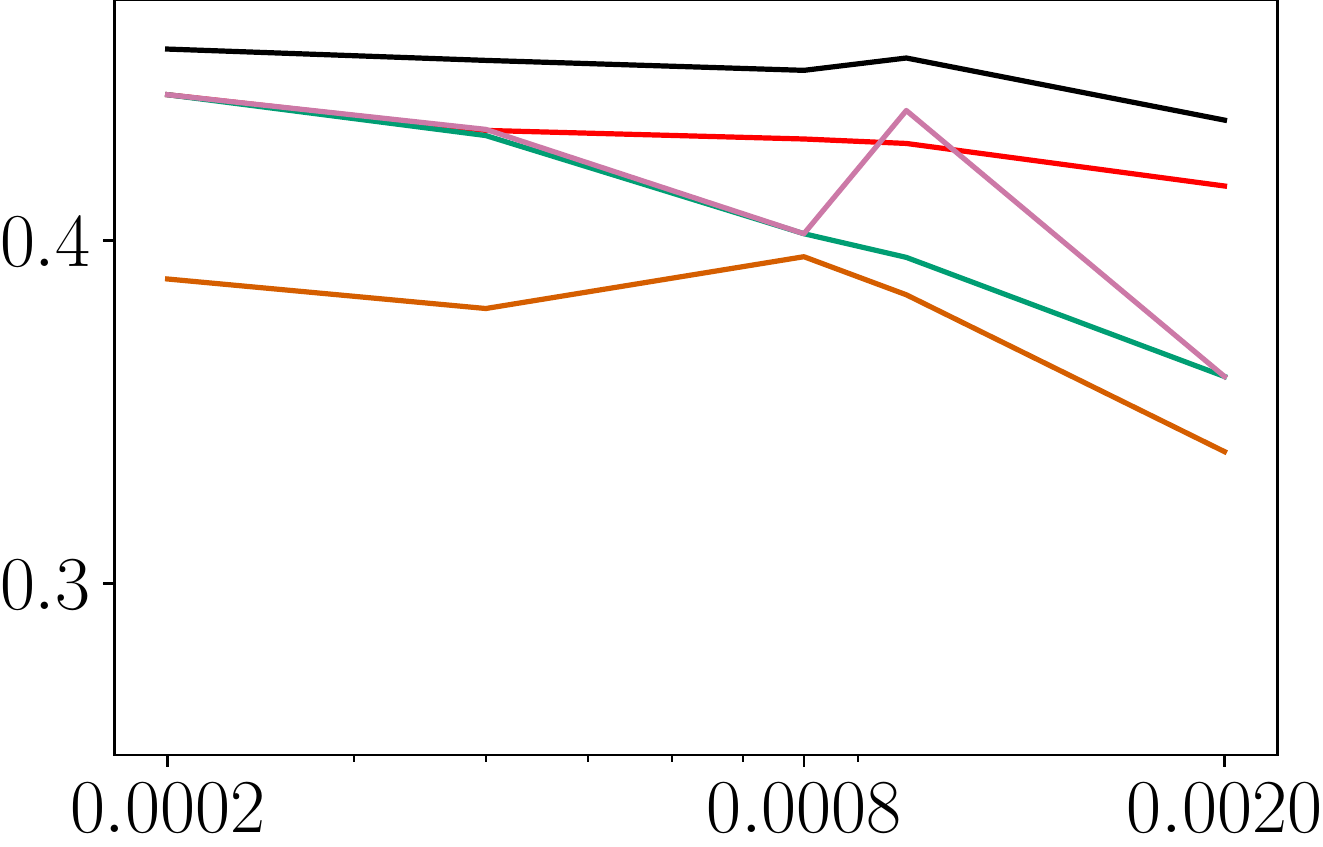}  \\
        & \small Learning rate
        & \small Learning rate
        & \small Learning rate
        
    \end{tabular}
    &
    \includegraphics[scale=0.5]{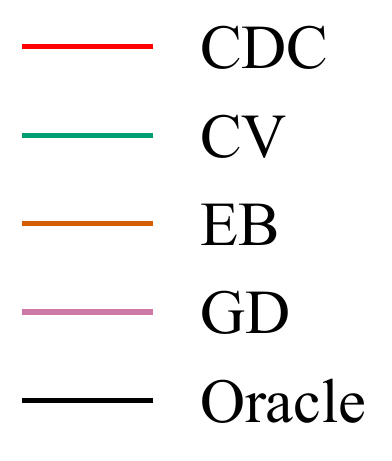}
    \end{tabular}
    \caption{\emph{\bfseries Linear models.} Performance comparison of \ourmethod with baselines in terms of test accuracy for linear models with different model sizes. Top: MNIST; Bottom: CIFAR-10; both with 50\% random labels.}
    \label{fig:generallinear}
\end{figure*}

{
\setlength{\tabcolsep}{0.1em}
\begin{figure}[t]
    \centering
    \begin{tabular}[]{lcc}
        & \small RMSE &  \small ListNet \\
        \rotatebox[origin=lt]{90}{\hspace{1em} \small NDCG@10}
        &
        \includegraphics[scale=0.26]{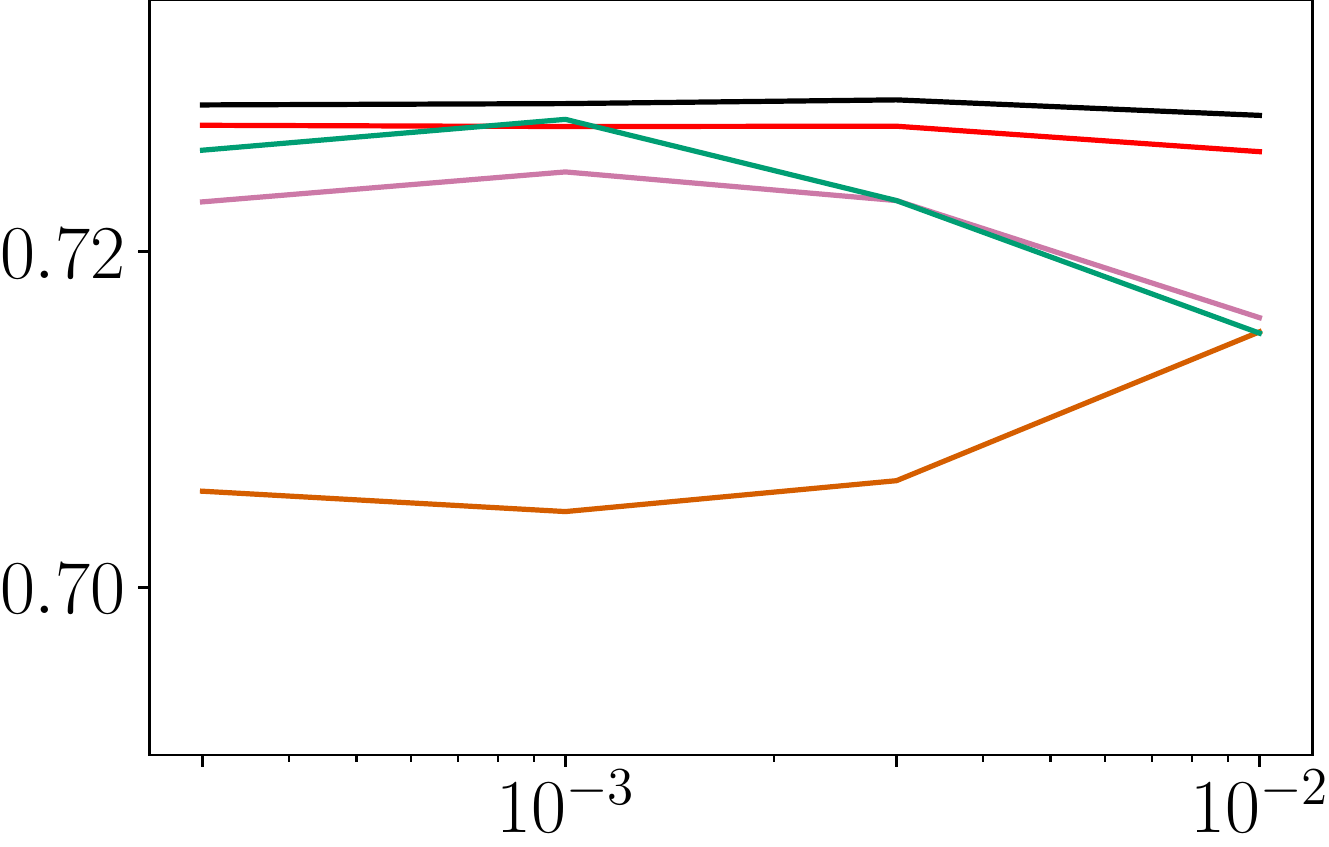}
        &
        \includegraphics[scale=0.26]{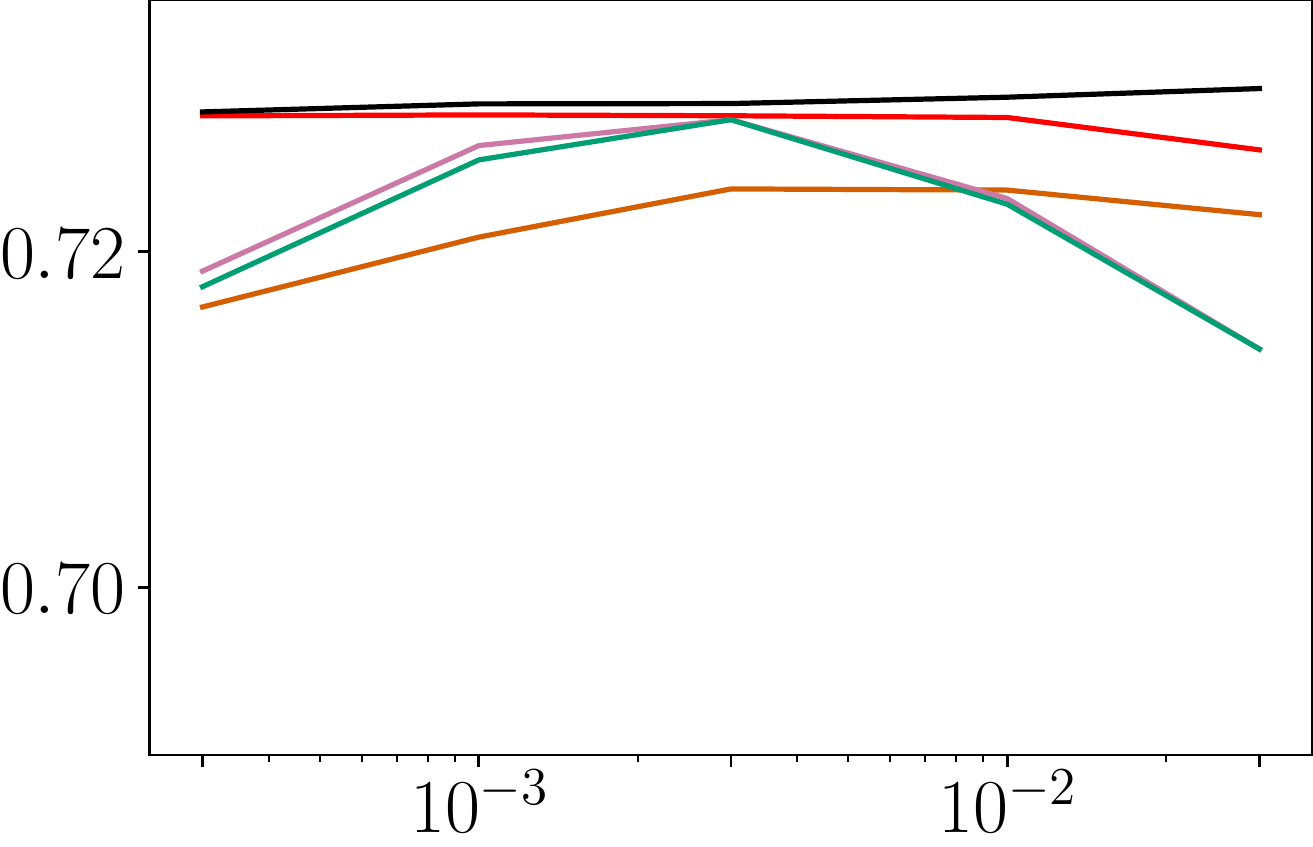}  \\
        \rotatebox[origin=lt]{90}{\hspace{1em} \small NDCG@10}
        &
        \includegraphics[scale=0.26]{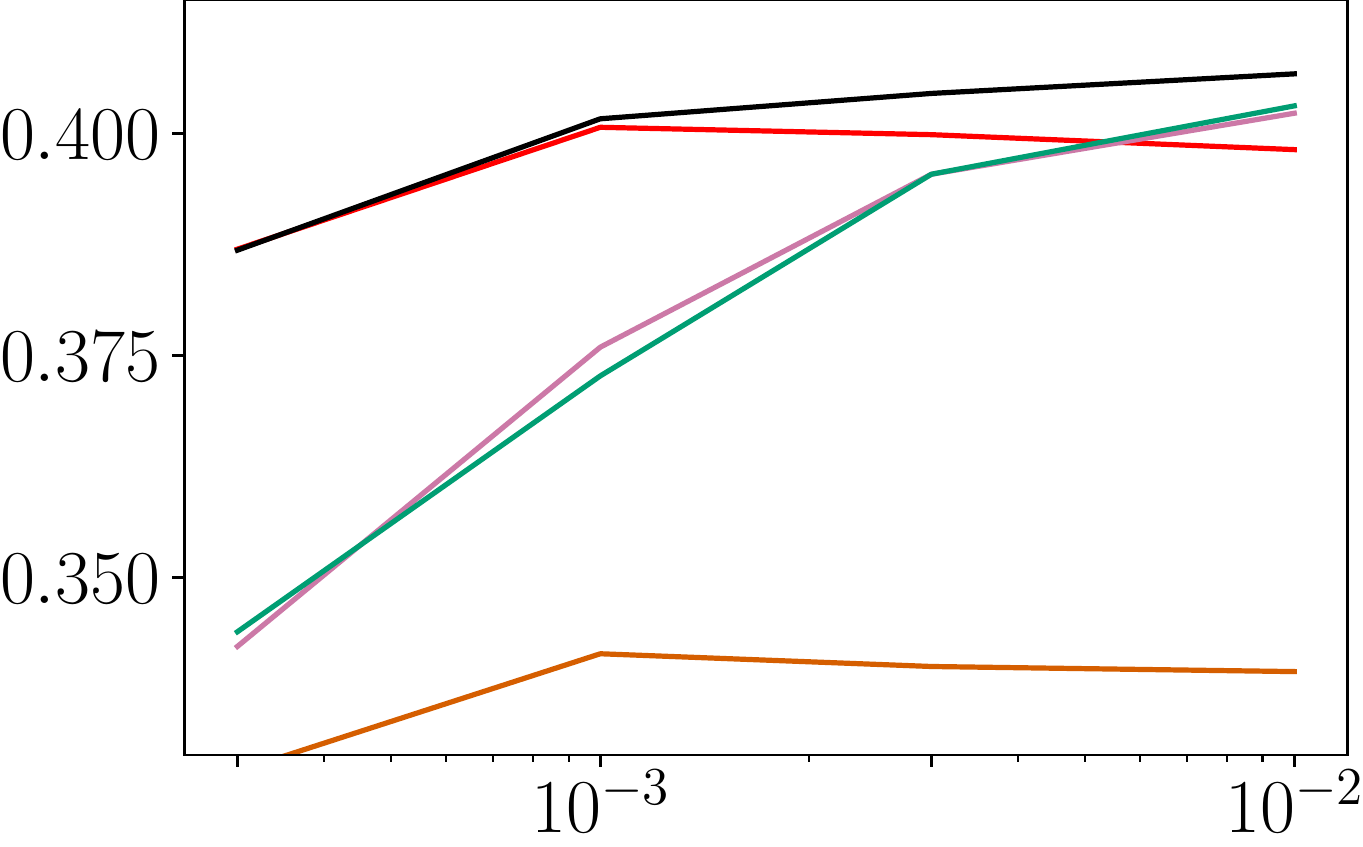}
        &
        \includegraphics[scale=0.26]{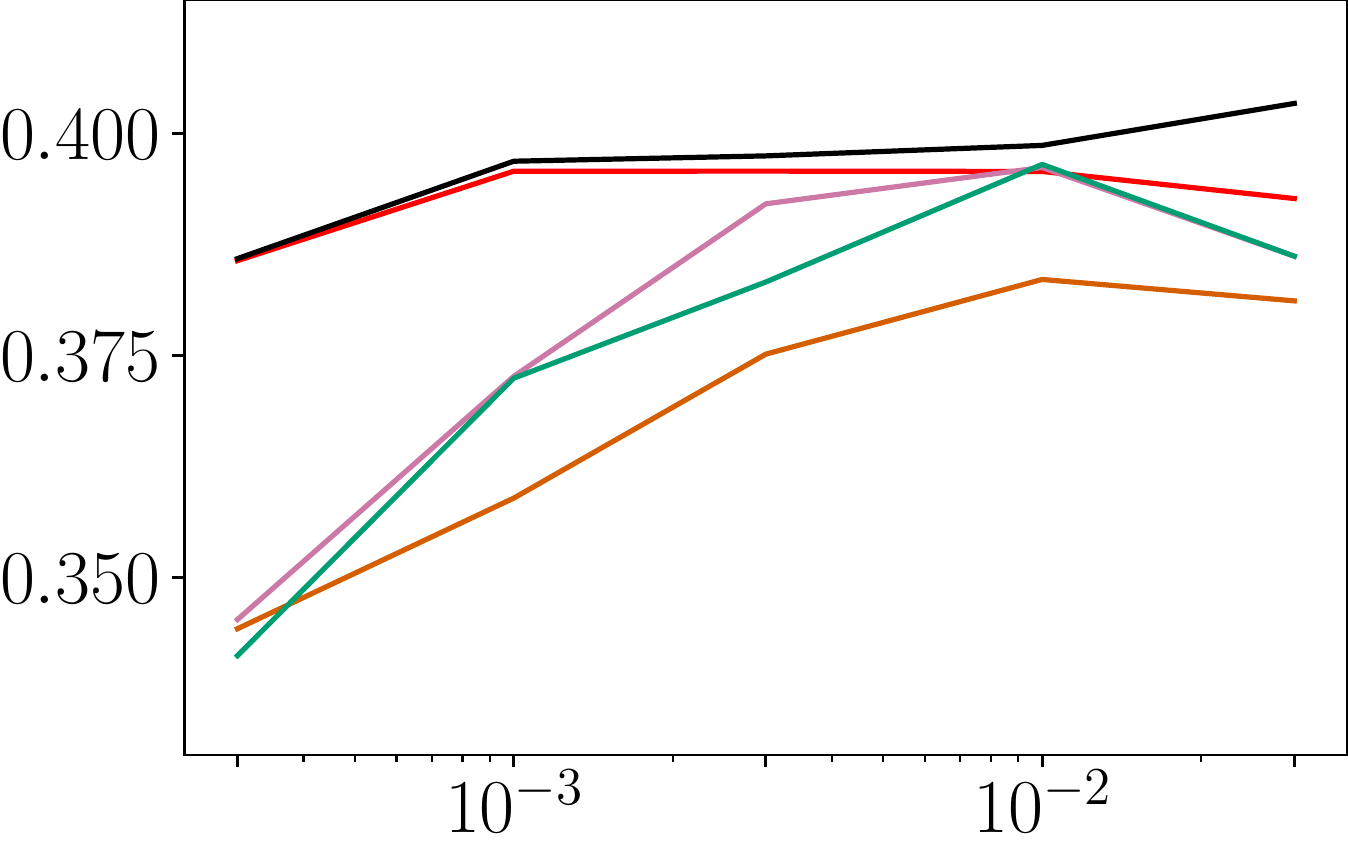} \\
        & \small Learning rate
        & \small Learning rate  \\
        \multicolumn{3}{c}{\includegraphics[scale=0.36]{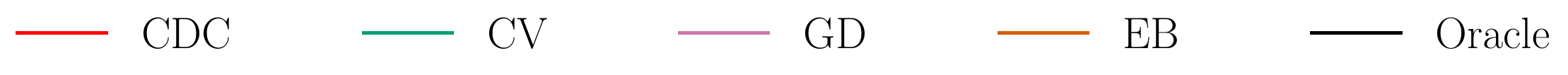}}
        
    \end{tabular}
    \caption{\emph{\bfseries Linear models.} Performance comparison of \ourmethod with baselines in terms of test NDCG@10 for different \ac{LTR} loss functions. Top: Yahoo! dataset; Bottom: MSLR dataset.}
    \label{fig:performance}
\end{figure}

}
\section{Results}
\label{sec:results}
In this section we present our experimental findings by comparing \ourmethod with other baselines (Sec.~\ref{subsec:baselines}) using both \moped linear and two-layer networks.


\subsection{Performance Comparison of Linear Models}
\label{subsec:performancecomparison} 
First, we show for a wide range of learning rates, that \ourmethod leads to near perfect generalization as opposed to other baselines.
Fig.~\ref{fig:generallinear} shows the performance comparison of \ourmethod with other baselines on image classification tasks for linear models with different sizes: small, medium and large as discussed in Sec.~\ref{sec:experiments}.
In each plot, we compare the generalization performance for a wide range of learning rates.
These plots show that, for almost all the learning rates, \ourmethod is better able to detect the start of overfitting and stop training, compared to other baselines.
Importantly, the gap between \ourmethod and Oracle is negligible in most cases, meaning that it is hard to improve our stopping rule for the tested cases.
As the learning rate is increased, degradation of test performance occurs more rapidly, making it more important and difficult to early stop at the correct iteration.
This is observed in these plots as the performance of different methods diverge for higher learning rates.
Even for this rather difficult task, we see \ourmethod{} performs robustly well and close to the Oracle.

Fig.~\ref{fig:performance} contains a performance comparison of \ourmethod with other baselines on the LTR datasets with RMSE and ListNet loss functions with a wide range of learning rates.
Here we observe similar results: \ourmethod is more robust in finding the correct generalizable point in all learning rates compared to other baselines and it performs very close to the skyline Oracle case.


A stopping rule that can have a performance very close to the oracle performance, reduces the sensitivity of the model to the choice of learning rate.
This means that
\begin{enumerate*}[label=(\roman*)]
    \item hyper-parameter tuning for learning rate would be of less concern; and
    \item relatively larger learning rates can be chosen for the model to converge faster; hence, significantly reducing the training time.
\end{enumerate*}

For a quantitative comparison between different stopping methods, in Table~\ref{tab:results:mean} we compare the mean and standard deviations of stopping methods across different setups (i.e., data points from Fig.~\ref{fig:generallinear},~\ref{fig:performance} and~\ref{fig:general2layer}).
This table shows that in five out of six cases, our \ourmethod performs significantly better than other methods on average.
It also has the least or second least variance for most of the cases.
These results further show the reliability of \ourmethod in different data modalities.

\subsection{Two-Layer Networks}
\label{sec:results:multilayer}
\begin{figure*}[t]
    \centering
    \begin{tabular}[]{cc}
    \begin{tabular}[]{lcccc}
        & \small Small & \small Medium & \small Large \\
        \rotatebox[origin=lt]{90}{\hspace{2em} \small Accuracy}
        &
        \includegraphics[scale=0.3]{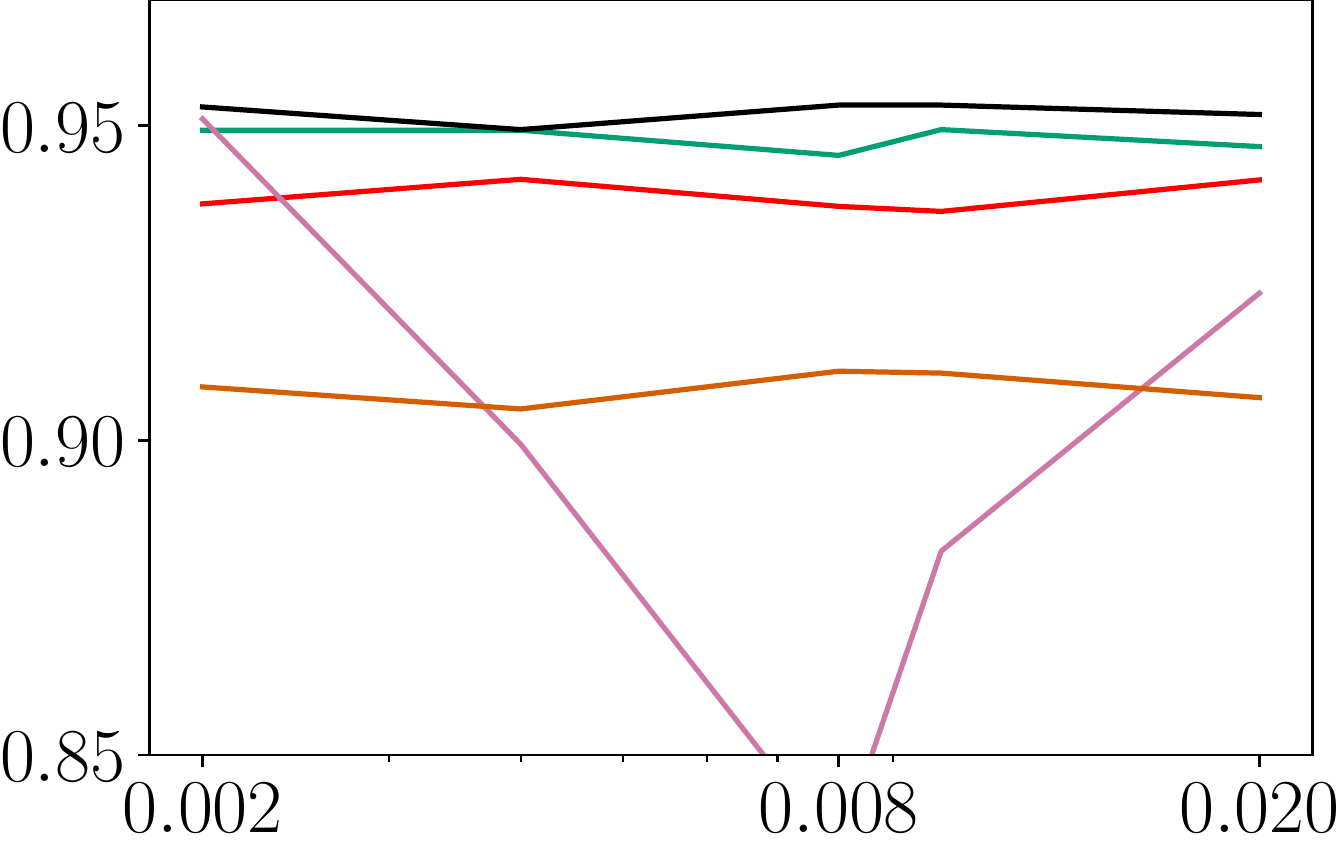}
        &
        \includegraphics[scale=0.3]{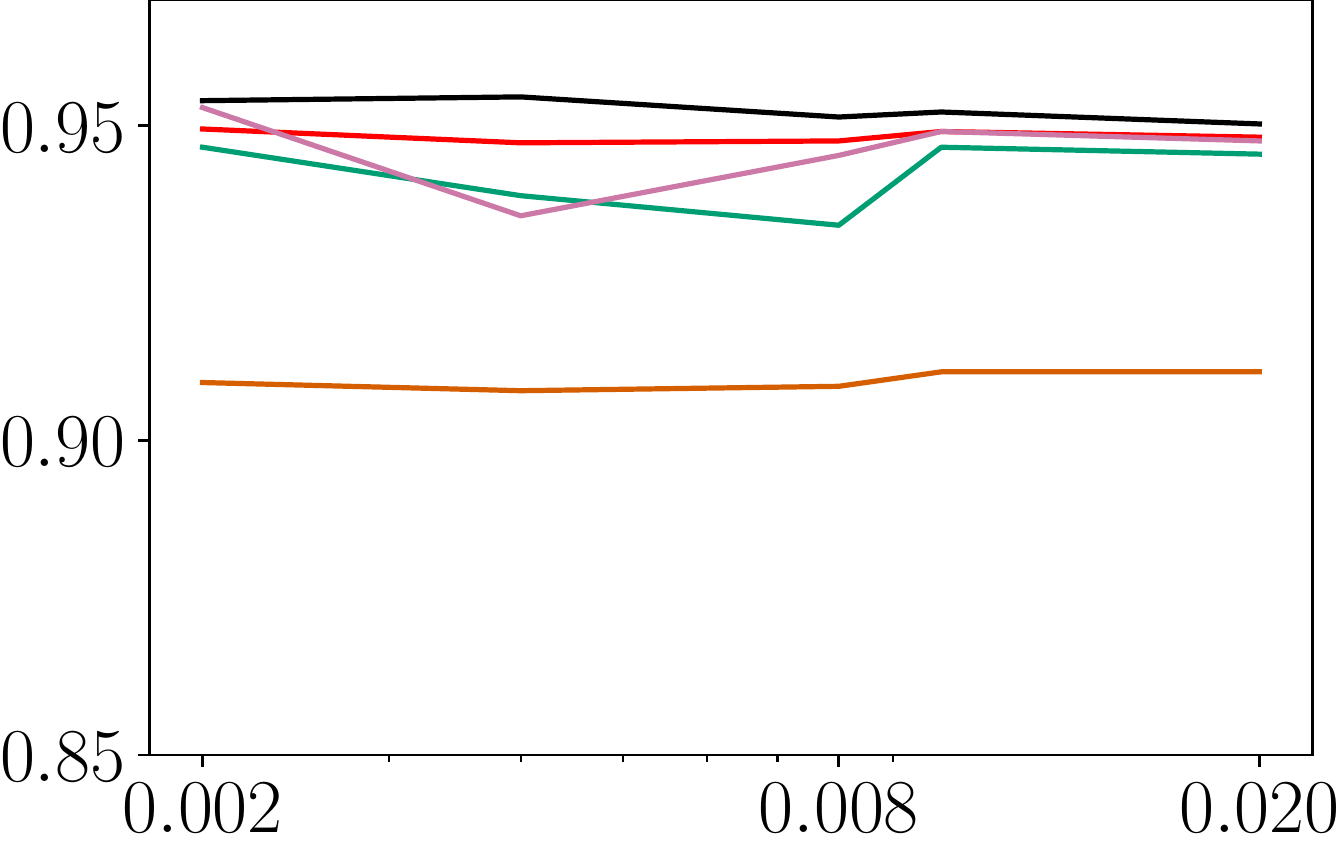}
        &
        \includegraphics[scale=0.3]{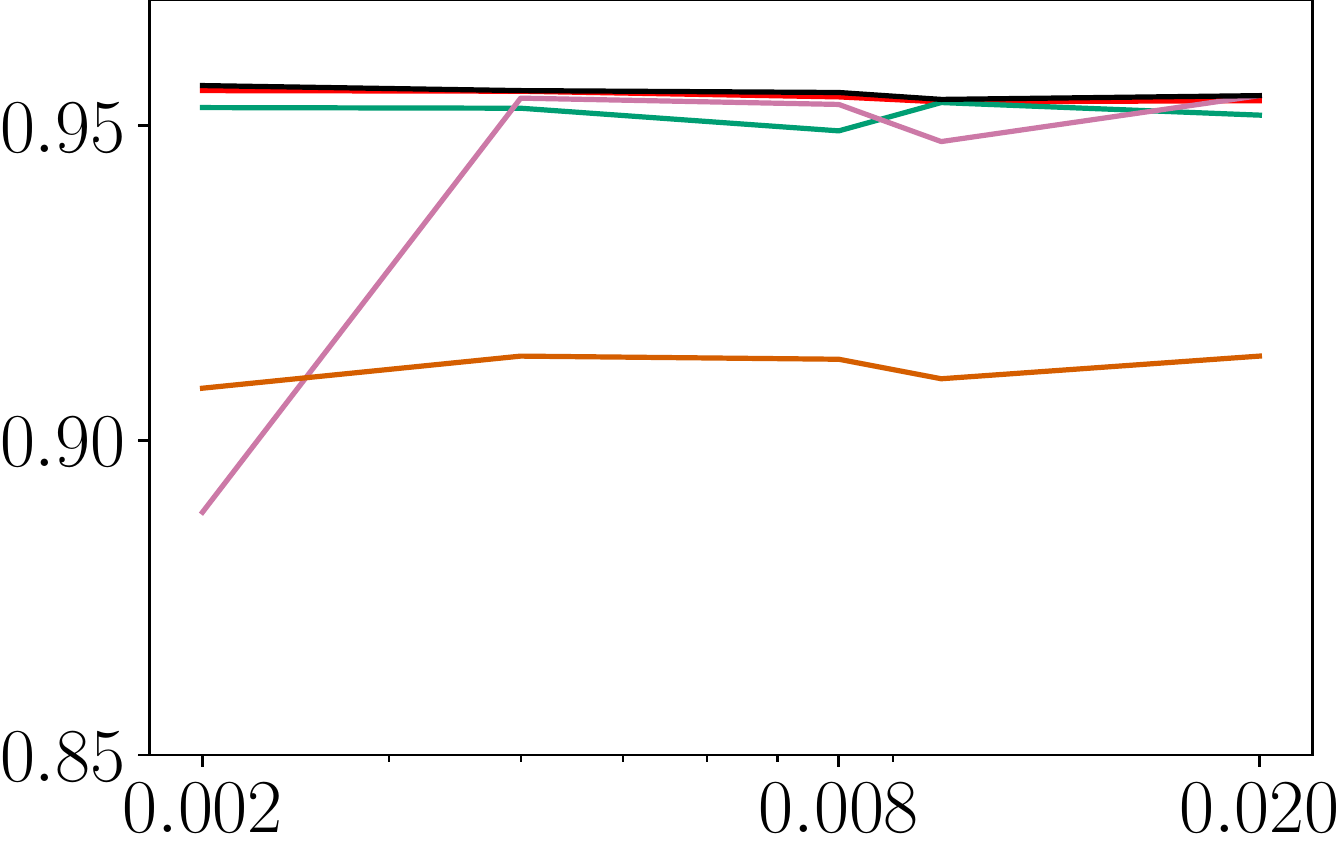}
        &
        \\
        \rotatebox[origin=lt]{90}{\hspace{2em} \small Accuracy}
        &
        \includegraphics[scale=0.3]{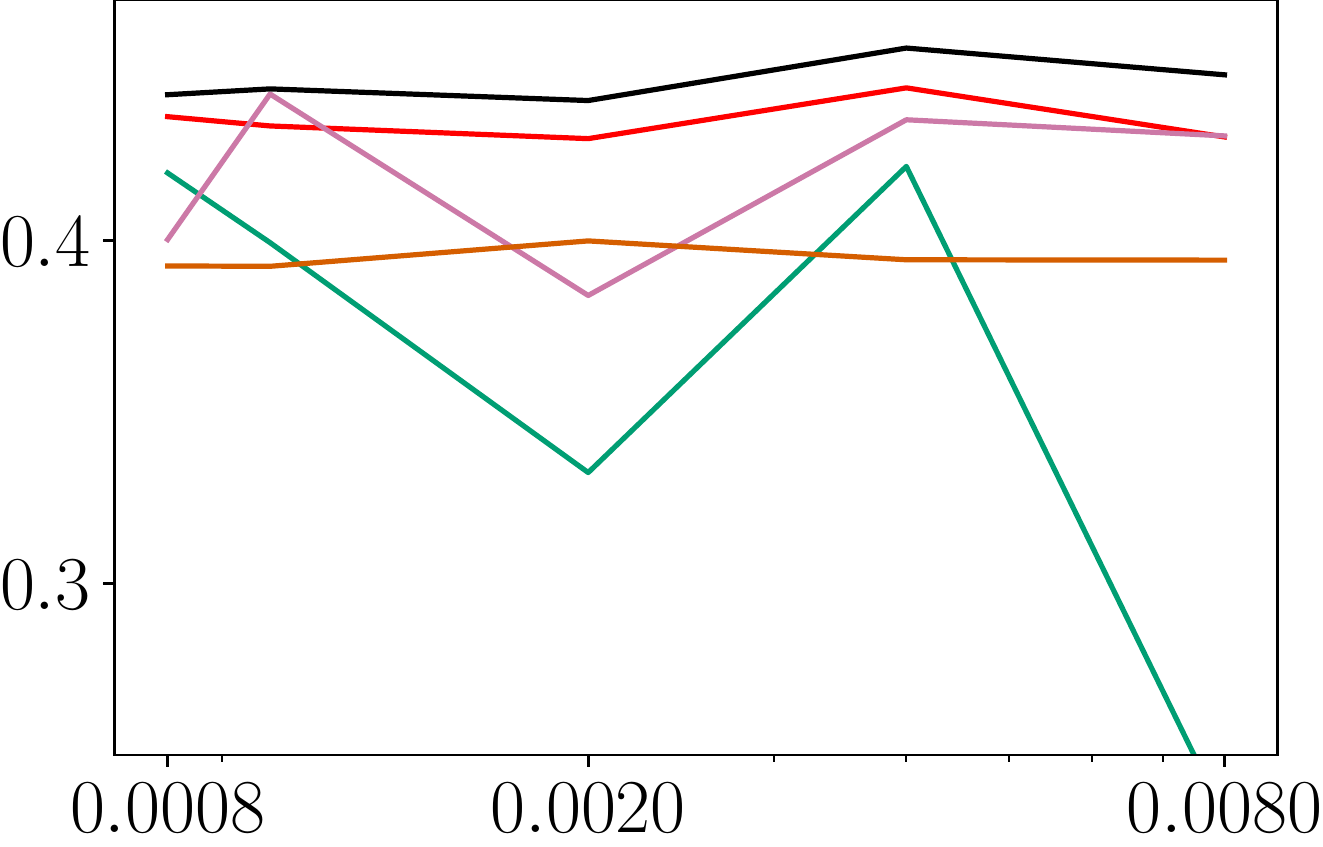}
        &
        \includegraphics[scale=0.3]{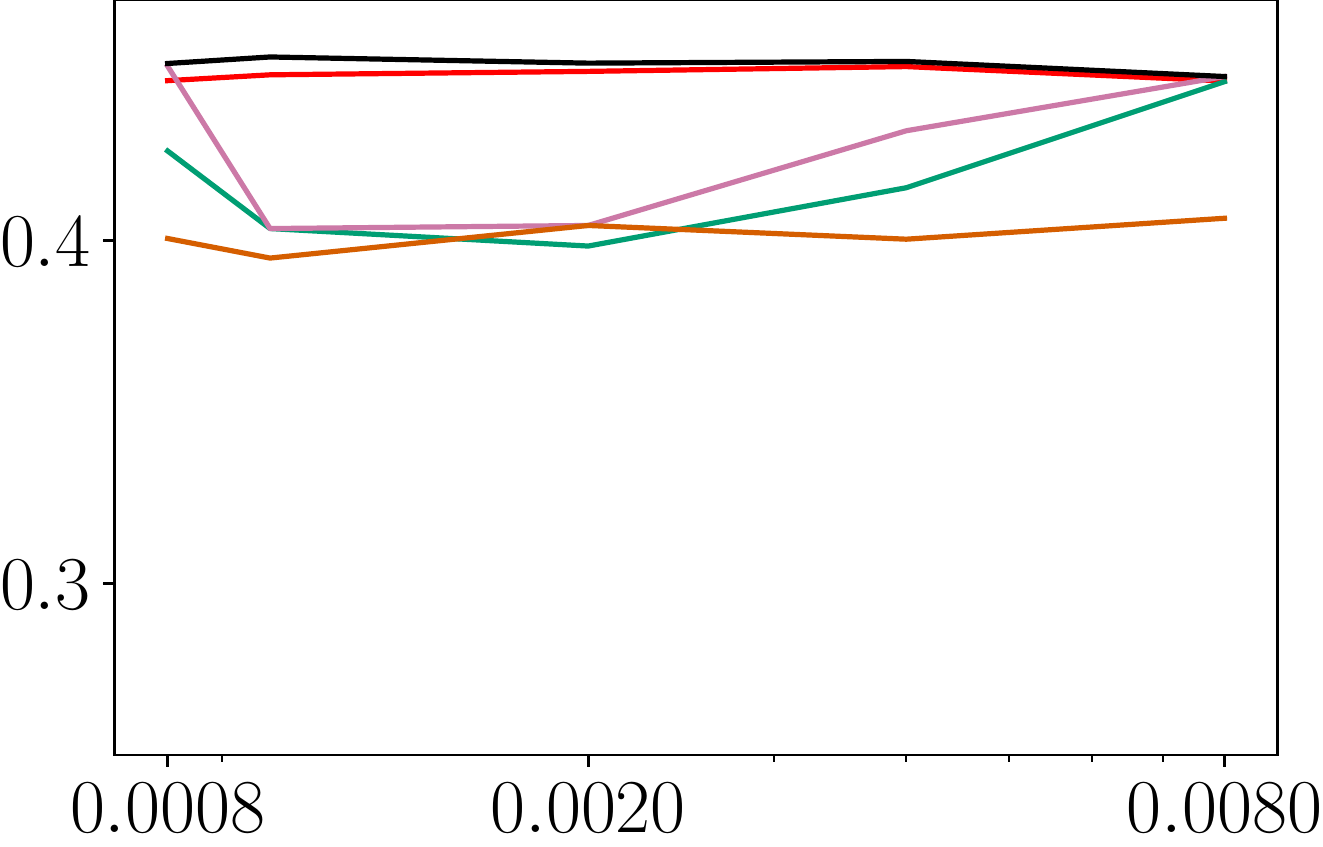}
        &
        \includegraphics[scale=0.3]{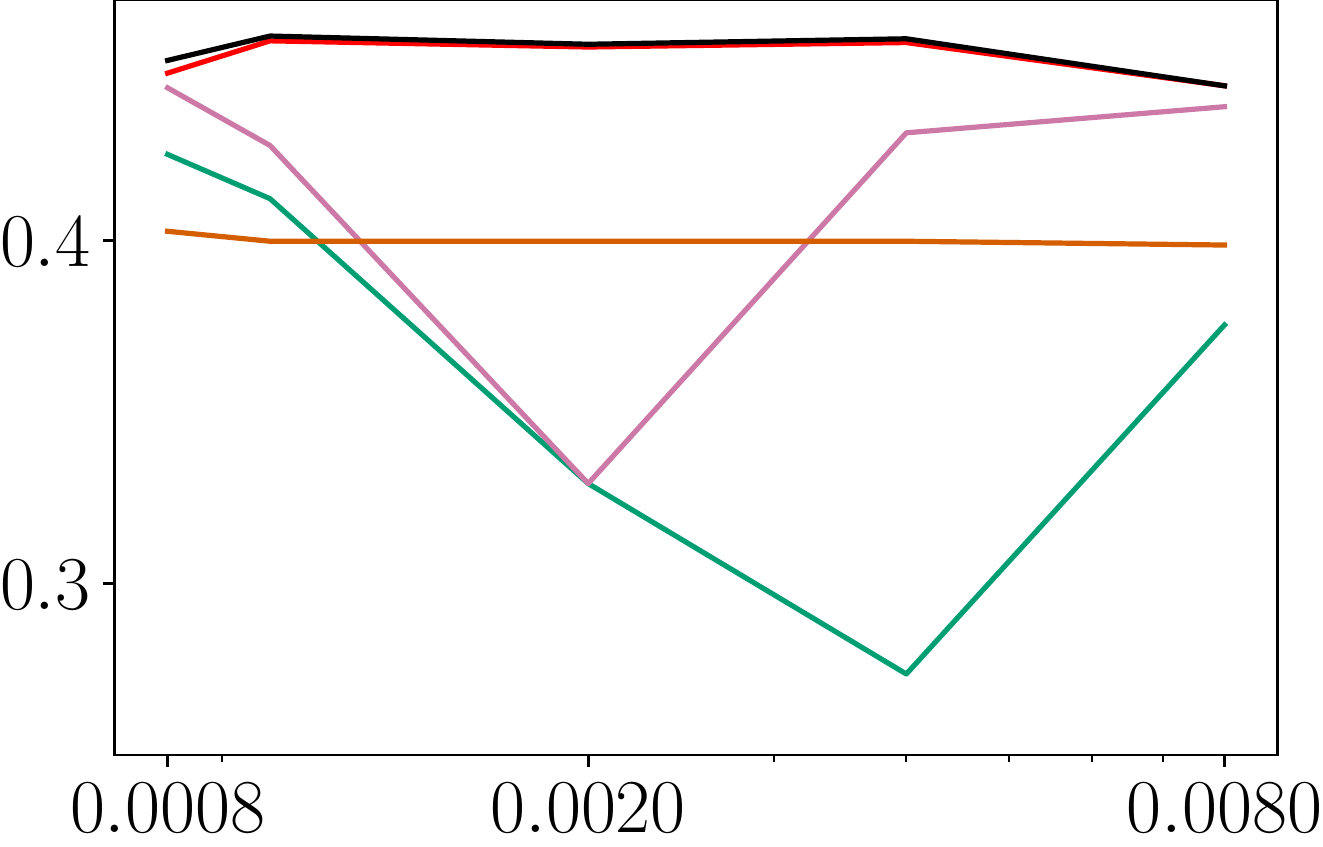}  \\
        & \small Learning rate
        & \small Learning rate
        & \small Learning rate
        
    \end{tabular}
    &
    \includegraphics[scale=0.5]{sections/figure/performance_legend_ver}
    \end{tabular}
    \caption{\emph{\bfseries Two-layer networks.} Performance comparison of \ourmethod with baselines in terms of test accuracy for two-layer networks with different model sizes. Top: MNIST; Bottom: CIFAR10; both with 50\% random labels.}
    \label{fig:general2layer}
\end{figure*}

Similar to results in the previous section on linear models, in this section the generalization performance of~\ourmethod with other baselines on two-layer networks is examined.
Fig.~\ref{fig:general2layer} contains the test performance of different baselines in terms of learning rates.
We observe similar results to the linear case here: \ourmethod performs more robustly on different learning rates compared to other baselines and it matches the Oracle performance, especially for medium and large networks.

The data points of these plots are aggregated and shown in Table~\ref{tab:results:mean} under columns indicated by ``2-layer''.
In this table we observe that for the MNIST dataset, \ourmethod and CV perform equally good and better than the other two baselines, while for the CIFAR-10 dataset, \ourmethod is the sole winner.

\begin{table*}[t]
    \caption{Mean and standard deviation of test accuracy and NDCG@10 across different model setups (Fig.~\ref{fig:generallinear},~\ref{fig:performance} and~\ref{fig:general2layer}).
    Bold and underlined entries indicate best and runner up values.
    Superscripts $^*$ indicate significance improvements over the next best score with $p<0.01$.
    }
    \label{tab:results:mean}
    \setlength{\tabcolsep}{0.6em}
    \begin{tabular}{l cccc cccc cccc}
    \toprule
        & \multicolumn{4}{c}{MNIST}
        & \multicolumn{4}{c}{CIFAR-10}
        & \multicolumn{2}{c}{Yahoo!}
        & \multicolumn{2}{c}{MSLR}
        \\
        \cmidrule(lr){2-5}\cmidrule(lr){6-9}\cmidrule(lr){10-11}\cmidrule(lr){12-13}
        & \multicolumn{2}{c}{Linear}
        & \multicolumn{2}{c}{2-layer}
        & \multicolumn{2}{c}{Linear}
        & \multicolumn{2}{c}{2-layer}
        & \multicolumn{2}{c}{Linear}
        & \multicolumn{2}{c}{Linear} \\
        
        \cmidrule(lr){2-3}\cmidrule(lr){4-5}\cmidrule(lr){6-7}\cmidrule(lr){8-9}\cmidrule(lr){10-11}\cmidrule(lr){12-13}
        &  Mean           & \small  \textit{STD}
        &  Mean           & \small  \textit{STD}
        &  Mean           & \small  \textit{STD}
        &  Mean           & \small  \textit{STD}
        &  Mean           & \small  \textit{STD}
        &  Mean           & \small  \textit{STD}
        \\
        \midrule
CDC & {\bfseries 0.939} & {\small \itshape \bfseries 0.0260} & {\underline{0.947}} & {\small \itshape 0.0069} & {\bfseries 0.431\rlap{$^*$}} & {\small \itshape \bfseries 0.0123} & {\bfseries 0.445\rlap{$^*$}} & {\small \itshape \underline{0.0093}} & {\bfseries 0.727\rlap{$^*$}} & {\small \itshape \bfseries 0.0035} & {\bfseries 0.394\rlap{$^*$}} & {\small \itshape \bfseries 0.0137}  \\
CV & {\underline{0.910}} & {\small \itshape 0.0586} & {\bfseries 0.947} & {\small \itshape \underline{0.0053}} & {0.396} & {\small \itshape 0.0380} & {0.381} & {\small \itshape 0.0618} & {\underline{0.722}} & {\small \itshape \underline{0.0067}} & {0.377} & {\small \itshape 0.0260}  \\
EB & {0.894} & {\small \itshape \underline{0.0321}} & {0.910} & {\small \itshape \bfseries 0.0024} & {0.379} & {\small \itshape \underline{0.0148}} & {0.399} & {\small \itshape \bfseries 0.0042} & {0.716} & {\small \itshape 0.0087} & {0.354} & {\small \itshape \underline{0.0231}}  \\
GD & {0.902} & {\small \itshape 0.0704} & {0.928} & {\small \itshape 0.0358} & {\underline{0.401}} & {\small \itshape 0.0370} & {\underline{0.420}} & {\small \itshape 0.0320} & {0.722} & {\small \itshape 0.0071} & {\underline{0.378}} & {\small \itshape 0.0241}  \\
\midrule
Oracle & {0.951} & {\small \itshape 0.0080} & {0.953} & {\small \itshape 0.0020} & {0.446} & {\small \itshape 0.0076} & {0.451} & {\small \itshape 0.0059} & {0.729} & {\small \itshape 0.0033} & {0.398} & {\small \itshape 0.0130}  \\
        \bottomrule
    \end{tabular}
    \end{table*}

\subsection{Discussion}
Our proposed method to find an early stopping point is based on detecting the intersection of two parallel instances of models.
To materialize such a detection, we use the cosine distance of the weight vectors of parallel instances.
This method of cosine distance works for linear models where all the learning is carried out by a single vector.
One use case of our criterion can be in linear probing as one of the popular methods for transferring a pre-trained model to a downstream task~\citep{peters2019tune, abnar2021exploring, kumar2022fine}.
With linear probing, we only train a linear layer that maps the representation from a frozen pre-trained backbone to the label space of the downstream task. Moreover, in general, in transfer learning, the downstream task has low training data. This lets \ourmethod{} to be a great choice for early stopping in linear probing setup and can lead to better generalization than the existing methods.

Going from one layer to more layers with non-linearities poses a challenge on how to effectively detect the intersection point.
Our proposed method in Sec.~\ref{sec:multilayer} considers the last layer of the instances to be able to continue to rely on the simple cosine distance.
Our extensive experiments confirm that by this extension our method will be as effective on two-layer networks as it is on linear models.
However, as the network gets deeper and more complex, the training dynamics might change and the last layer may not be sufficient to detect intersection of parallel instances.
Consequently, our cosine distance criterion works best for linear and two-layer networks, and possibly for linear probing a pre-trained model.
We believe that the general idea of this paper, i.e., using the intersection of two parallel instances of a model to signal early stopping, can be applied to more complex models, but with a more complex criterion that involves the whole model and not just the last layer.
We leave further development of more complex intersection detection methods for future work.


\section{Conclusion and Future Work}
\label{sec:conclusion} 
We have proposed the \ourmethodtitle for early stopping in linear and two-layer networks.
Our stopping rule does not depend on a separate validation, nor does it use cross validation.
This means the entire labeled data can be used for training.
\ourmethod is based on the consistent observation that two parallel instances of a linear model, initialized with different random seeds, will converge to the same solution on the fitness surface.
This observation, together with supporting evidence from~\citep{li2020gradient} on the distance of overfit weights to the initialized values of a network, led us to propose an early stopping rule based on the cosine distance of two parallel instances of an \moped linear model.
The intuition is that when two randomly initialized weights start to become aligned, the parallel instances will intersect, and this is a signal for overfitting.

We have compared the generalization of our \ourmethod{} rule with existing methods, namely cross validation, evidence-based~\citep{mahsereci2017early} and gradient disparity~\citep{forouzesh2021disparity}, and shown that in all of the tested datasets and all model setups, \ourmethod performs more reliably with different learning rates and there is a small gap between its performance and the skyline Oracle performance.
We have also extended our method to work with multi-layer networks, using our notion of counterfactual weights vector.
We have shown theoretically that our proposal is equivalent to comparing the projection of the output vectors of two instances, using a special projection matrix.
Importantly, we have argued why the output vectors themselves cannot be used to detect the intersection point and verified our arguments experimentally.\looseness=-1

The most interesting future direction to this work is to verify our conjecture about the intersection of parallel instances as the start of overfitting.
Furthermore, finding a more complex criterion than the cosine distance to detect this intersection that involves the whole model and not just the last layer, and is suitable for more complex structures as well, would be a natural follow up to this work.\looseness=-1

\section*{Code and data}
To facilitate the reproducibility of the reported results, this work only made use of publicly available data and our experimental implementation is publicly available at \url{https://github.com/AliVard/CDC-Early-Stopping}.

\begin{acks}
This research was supported by Elsevier and the Netherlands Organisation for Scientific Research (NWO)
under pro\-ject nr
612.\-001.\-551, and
by the Hybrid Intelligence Center, a 10-year program funded by the Dutch Ministry of Education, Culture and Science through the Netherlands Organisation for Scientific Research, \url{https://hybrid-intelligence-centre.nl}.

All content represents the opinion of the authors, which is not necessarily shared or endorsed by their respective employers and/or sponsors.\looseness=-1
\end{acks}
 
\clearpage
\bibliographystyle{ACM-Reference-Format}
\balance
\bibliography{references}

\end{document}